\documentclass{article}
\usepackage{}
\usepackage{amsmath, algorithm, graphicx, subcaption, siunitx, float, hyperref, cleveref, amssymb}
\usepackage[noend]{algpseudocode}
\usepackage[utf8]{inputenc}
\usepackage{booktabs}
\usepackage[blocks]{authblk}

\makeatletter
\def\BState{\State\hskip-\ALG@thistlm}
\makeatother

\hypersetup{
    colorlinks=true,
    linkcolor=black,
    citecolor=black,
    urlcolor=blue,
}

\usepackage{natbib}
\setcitestyle{authoryear,open={(},close={)}}

\newcommand{\changeurlcolor}[1]{\hypersetup{urlcolor=#1}} 

\nocite{*}

\title{{\Large Let's Verify Step by Step}}

\newcommand\CoAuthorMark{\footnotemark[\arabic{footnote}]}
\author{
    \textbf{\scriptsize Hunter Lightman\footnote{Primary authors. Correspondence to: Karl Cobbe \textless karl@openai.com\textgreater}\hspace{6mm}}
    \textbf{\scriptsize Vineet Kosaraju\protect\CoAuthorMark \hspace{6mm}}
    \textbf{\scriptsize Yura Burda\protect\CoAuthorMark \hspace{6mm}}
    \textbf{\scriptsize Harri Edwards\hspace{5mm}}\\
    \vspace{.1cm}
    \textbf{\scriptsize Bowen Baker\hspace{5mm}}
    \textbf{\scriptsize Teddy Lee\hspace{5mm}}
    \textbf{\scriptsize Jan Leike\hspace{5mm}}
    \textbf{\scriptsize John Schulman\hspace{5mm}}
    \textbf{\scriptsize Ilya Sutskever\hspace{5mm}}\\
    \vspace{.1cm}
    \textbf{\scriptsize Karl Cobbe\protect\CoAuthorMark}
}

\affil{\small OpenAI}
\date{}

\begin{document}

\maketitle

\vspace{-.5cm}
\begin{abstract}

In recent years, large language models have greatly improved in their ability to perform complex multi-step reasoning. However, even state-of-the-art models still regularly produce logical mistakes. To train more reliable models, we can turn either to outcome supervision, which provides feedback for a final result, or process supervision, which provides feedback for each intermediate reasoning step. Given the importance of training reliable models, and given the high cost of human feedback, it is important to carefully compare the both methods. Recent work has already begun this comparison, but many questions still remain. We conduct our own investigation, finding that process supervision significantly outperforms outcome supervision for training models to solve problems from the challenging MATH dataset. Our process-supervised model solves 78\% of problems from a representative subset of the MATH test set. Additionally, we show that active learning significantly improves the efficacy of process supervision. To support related research, we also release PRM800K, the complete dataset of 800,000 step-level human feedback labels used to train our best reward model.

\end{abstract}

\section{Introduction}

Large language models are capable of solving tasks that require complex multi-step reasoning by generating solutions in a step-by-step chain-of-thought format \citep{nye2021show, wei2022chain, kojima2022large}. However, even state-of-the-art models are prone to producing falsehoods --- they exhibit a tendency to invent facts in moments of uncertainty \citep{bubeck2023sparks}. These \textit{hallucinations} \citep{maynez2020faithfulness} are particularly problematic in domains that require multi-step reasoning, since a single logical error is enough to derail a much larger solution. Detecting and mitigating hallucinations is essential to improve reasoning capabilities.

One effective method involves training reward models to discriminate between desirable and undesirable outputs. The reward model can then be used in a reinforcement learning pipeline \citep{ziegler2019fine, stiennon2020learning, nakano2021webgpt, ouyang2022training} or to perform search via rejection sampling \citep{nichols2020collaborative, shen2021generate, cobbe2021training}. While these techniques are useful, the resulting system is only as reliable as the reward model itself. It is therefore important that we study how to most effectively train reliable reward models.

In closely related work, \cite{uesato2022solving} describe two distinct methods for training reward models: outcome supervision and process supervision. Outcome-supervised reward models (ORMs) are trained using only the final result of the model's chain-of-thought, while process-supervised reward models (PRMs) receive feedback for each step in the chain-of-thought. There are compelling reasons to favor process supervision. It provides more precise feedback, since it specifies the exact location of any errors that occur. It also has several advantages relevant to AI alignment: it is easier for humans to interpret, and it more directly rewards models for following a human-endorsed chain-of-thought. Within the domain of logical reasoning, models trained with outcome supervision regularly use incorrect reasoning to reach the correct final answer \citep{zelikman2022star, creswell2022selection}. Process supervision has been shown to mitigate this misaligned behavior \citep{uesato2022solving}.

Despite these advantages, \cite{uesato2022solving} found that outcome supervision and process supervision led to similar final performance in the domain of grade school math. We conduct our own detailed comparison of outcome and process supervision, with three main differences: we use a more capable base model, we use significantly more human feedback, and we train and test on the more challenging MATH dataset \citep{hendrycks2021measuring}.

Our main contributions are as follows:

\begin{enumerate}
\item We show that process supervision can train much more reliable reward models than outcome supervision. We use our state-of-the-art PRM to solve $78.2\%$ of problems from a representative subset of the MATH test set.

\item We show that a large reward model can reliably approximate human supervision for smaller reward models, and that it can be used to efficiently conduct large-scale data collection ablations.

\item We show that active learning leads to a $2.6\times$ improvement in the data efficiency of process supervision.

\item We release our full process supervision dataset, PRM800K, to promote related research.

\end{enumerate}

\section{Methods}

We perform a comparison of outcome and process supervision, following a similar methodology to \cite{uesato2022solving}. Outcome supervision can be provided without humans, since all problems in the MATH dataset have automatically checkable answers. In contrast, there is no simple way to automate process supervision. We therefore rely on human data-labelers to provide process supervision, specifically by labelling the correctness of each step in model-generated solutions.

We conduct experiments in two separate regimes: large-scale and small-scale. Each has its own advantages, and they offer complimentary perspectives. At large-scale, we finetune all models from GPT-4 \citep{gpt4}. We focus on advancing the state-of-the-art by training the most reliable ORM and PRM possible. Unfortunately the training sets for these reward models are not directly comparable, for reasons we will discuss in \Cref{section:large_scale}. These models are therefore not ideal for making an apples-to-apples comparison of outcome and process supervision. To address this flaw, we also train models at small-scale, where we can conduct a more direct comparison. In order to remove our dependence on costly human feedback, we use a large-scale model to supervise small-scale model training. This setup enables us to conduct several important ablations that would otherwise be infeasible.

\subsection{Scope}

At each model scale, we use a single fixed model to generate all solutions. We call this model the \textit{generator}. We do not attempt to improve the generator with reinforcement learning (RL). When we discuss outcome and process supervision, we are specifically referring to the supervision given to the reward model. We do not discuss any supervision the generator would receive from the reward model if trained with RL. Although finetuning the generator with RL is a natural next step, it is intentionally not the focus of this work.

We instead focus exclusively on how to train the most reliable reward model possible. We evaluate a reward model by its ability to perform best-of-N search over uniformly sampled solutions from the generator. For each test problem we select the solution ranked highest by the reward model, automatically grade it based on its final answer, and report the fraction that are correct. A reward model that is more reliable will select the correct solution more often.

\subsection{Base Models} \label{section:base_models}

All large-scale models are finetuned from the base GPT-4 model \citep{gpt4}. This model has been pretrained solely to predict the next token; it has not been pretrained with any Reinforcement Learning from Human Feedback (RLHF) \citep{christiano2017deep}. The small-scale base models are similar in design to GPT-4, but they were pretrained with roughly $200$ times less compute. As an additional pretraining step, we finetune all models on a dataset of roughly 1.5B math-relevant tokens, which we call MathMix. Similar to \cite{lewkowycz2022solving}, we find that this improves the model's mathematical reasoning capabilities. Details on how this dataset was constructed can be found in \Cref{appendix:mathmix}.

\subsection{Generator} \label{section:generator}

To make parsing individual steps easier, we train the generator to produce solutions in a newline delimited step-by-step format. Specifically, we few-shot generate solutions to MATH training problems, filter to those that reach the correct final answer, and finetune the base model on this dataset for a single epoch. This step is not intended to teach the generator new skills; it is intended only to teach the generator to produce solutions in the desired format. 

\begin{figure}
\centering
\begin{subfigure}{\textwidth}
\includegraphics[width=\textwidth]{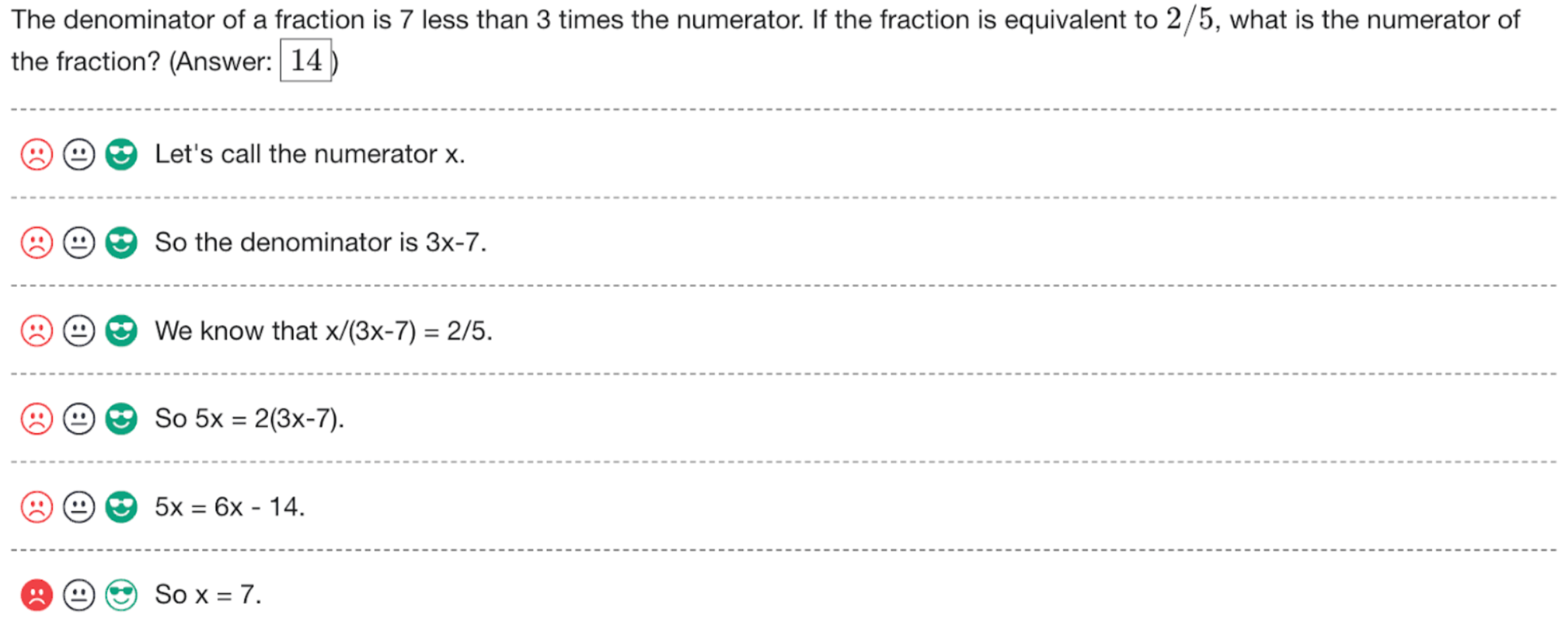}
\end{subfigure}
\caption{A screenshot of the interface used to collect feedback for each step in a solution.}
\label{figure:data_interface}
\end{figure}

\subsection{Data Collection} \label{section:data_collection}

To collect process supervision data, we present human data-labelers with step-by-step solutions to MATH problems sampled by the large-scale generator. Their task is to assign each step in the solution a label of \textit{positive}, \textit{negative}, or \textit{neutral}, as shown in \Cref{figure:data_interface}. A positive label indicates that the step is correct and reasonable. A negative label indicates that the step is either incorrect or unreasonable. A neutral label indicates ambiguity. In practice, a step may be labelled neutral if it is subtly misleading, or if it is a poor suggestion that is technically still valid. We permit neutral labels since this allows us to defer the decision about how to handle ambiguity: at test time, we can treat neutral labels as either positive or negative. A more detailed description of the labelling instructions is provided in \Cref{appendix:labelling_instructions}.

We label solutions exclusively from the large-scale generator in order to maximize the value of our limited human-data resource. We refer to the entire dataset of step-level labels collected as PRM800K. The PRM800K training set contains 800K step-level labels across 75K solutions to 12K problems. To minimize overfitting, we include data from 4.5K MATH test problems in the PRM800K training set, and we therefore evaluate our models only on the remaining $500$ MATH test problems. More details about this test set can be found in \Cref{appendix:evaluation}.

During data collection, we must decide which solutions to surface to data-labelers. The most straightforward strategy is to uniformly surface solutions produced by the generator. However, if we surface solutions that make obvious errors, the human feedback we get is less valuable. We would prefer to surface solutions that are more likely to fool our best reward model. To that end, we attempt to strategically select which solutions to show data-labelers. Specifically, we choose to surface \textit{convincing wrong-answer} solutions. We use the term \textit{convincing} to refer to solutions that are rated highly by our current best PRM, and we use \textit{wrong-answer} to refer to solutions that reach an incorrect final answer. We use this slightly verbose phrasing to emphasize the fact that correctness is determined solely by checking the final answer, a process which occasionally leads to misgraded solutions. We expect to gain more information from labeling convincing wrong-answer solutions, since we know the PRM is mistaken about at least one step in each such solution.

In addition to using this selection strategy, we also iteratively re-train our PRM using the latest data at several points in the data collection process. At each iteration, we generate N solutions per problem and surface only the top K most convincing wrong-answer solutions to data-labelers. We experiment with either applying this top-K filtering at a problem level (K solutions per problem) or globally across the dataset (K solutions in total, unequally distributed among problems). Since the data collection process is expensive, it was not feasible to conduct at-scale ablations of these decisions. However, we perform several surrogate ablations in \Cref{section:synthetic_supervision}, using our largest PRM as a labelling oracle for a smaller PRM. More details about data collection can be found in \Cref{appendix:data_collection}.

\begin{figure}
\centering
\begin{subfigure}[t]{0.49 \textwidth}
\vskip 0pt
\includegraphics[width=\textwidth]{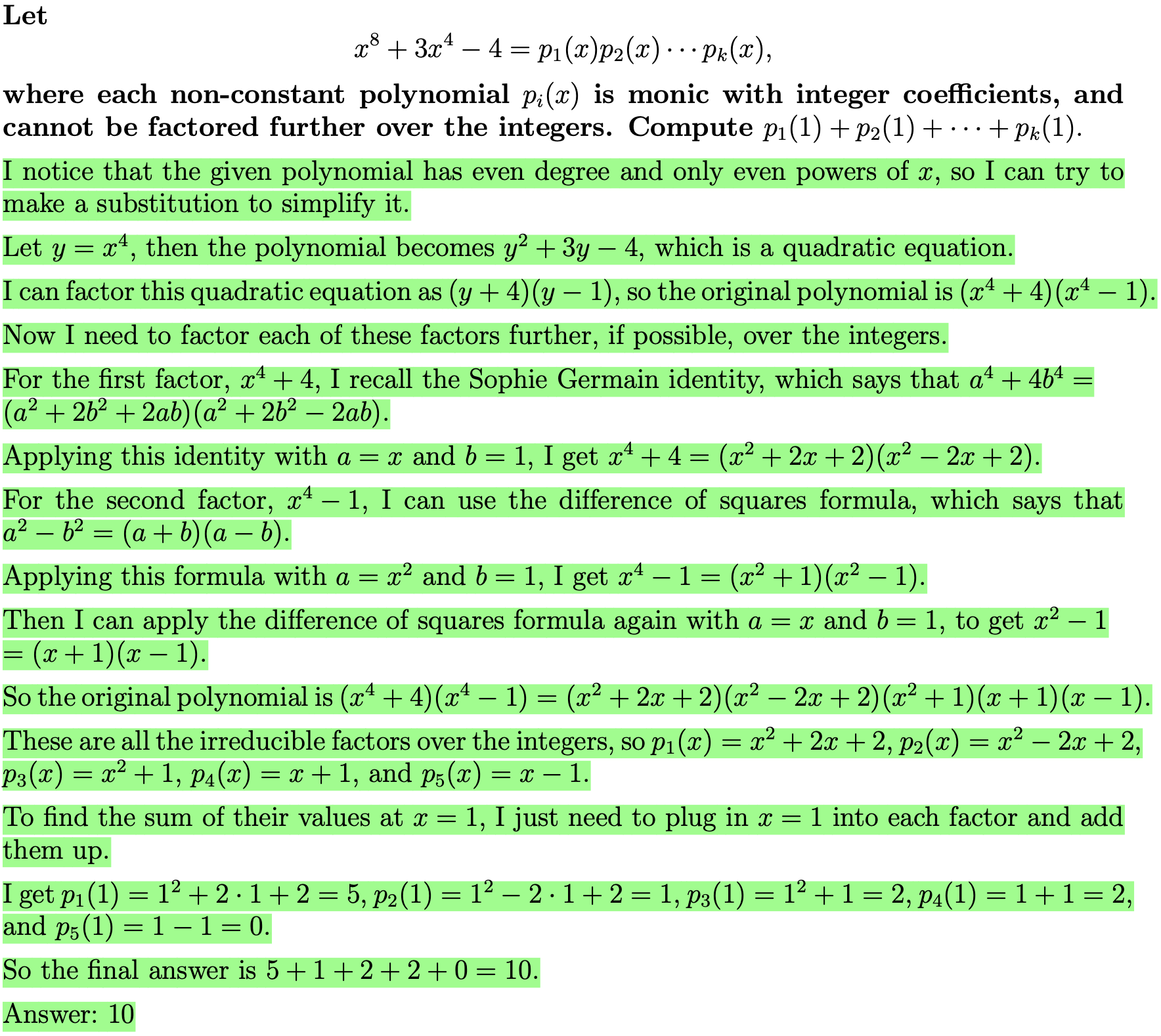}
\end{subfigure}
\begin{subfigure}[t]{0.49 \textwidth}
\vskip 0pt
\includegraphics[width=\textwidth]{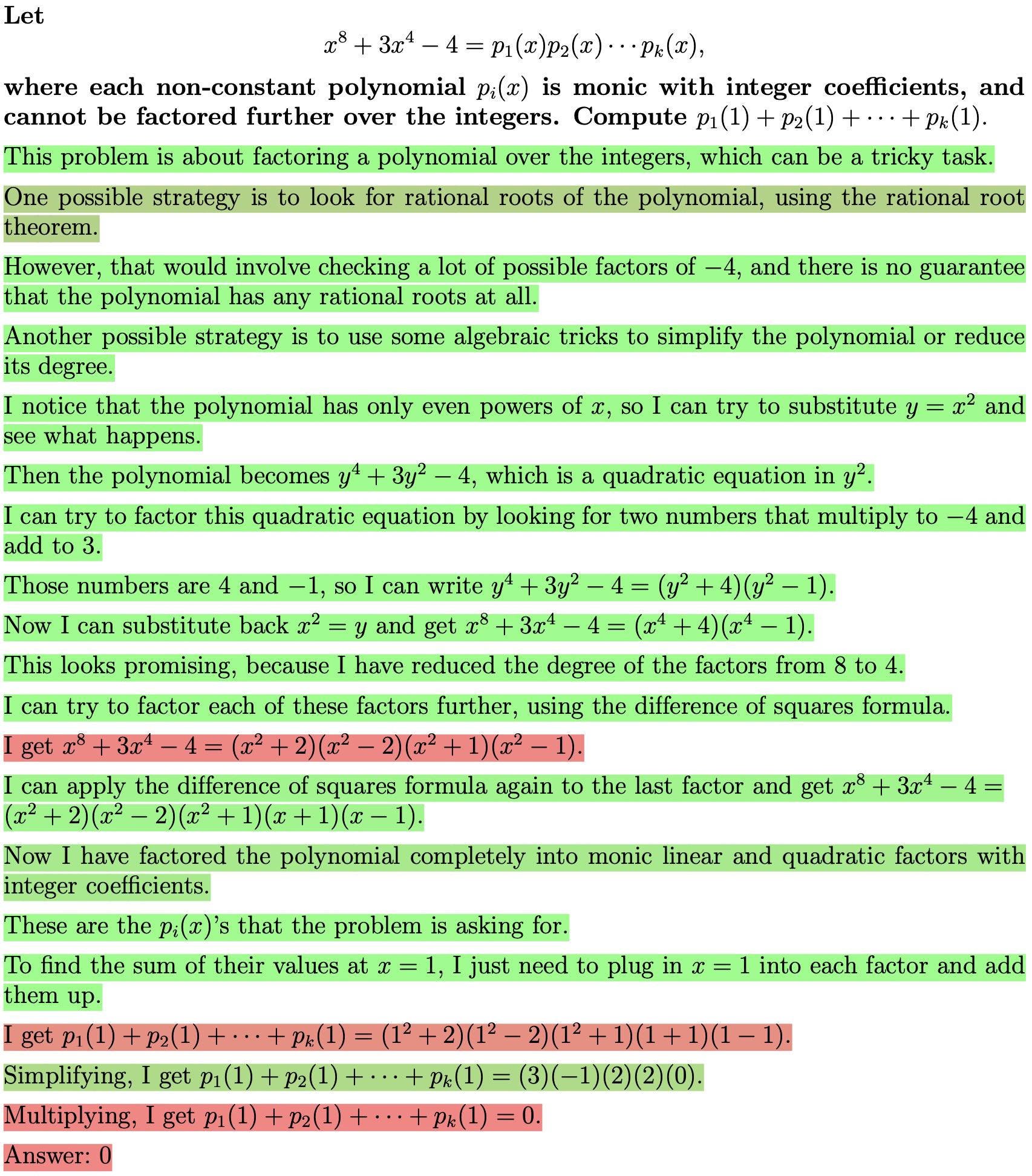}
\end{subfigure}
\caption{Two solutions to the same problem, graded by the PRM. The solution on the left is correct while the solution on the right is incorrect. A green background indicates a
high PRM score, and a red background indicates a low score. The PRM correctly identifies the mistake in the incorrect solution.}
\label{figure:prm_grading_examples}
\end{figure}

\subsection{Outcome-supervised Reward Models (ORMs)} \label{section:orm}

We train ORMs following a similar methodology to \cite{cobbe2021training}. We uniformly sample a fixed number of solutions per problem from the generator, and we train the ORM to predict whether each solution is correct or incorrect. In practice, we usually determine correctness by automatically checking the final answer, but in principle these labels could be provided by humans. At test time, we use the ORM's prediction at the final token as the overall score for the solution. We note the automatic grading used to determine ORM targets is not perfectly reliable: \textit{false positives} solutions that reach the correct answer with incorrect reasoning will be misgraded. We discuss additional ORM training details in \Cref{appendix:orm_details}.

\subsection{Process-supervised Reward Models (PRMs)}

We train PRMs to predict the correctness of each step after the last token in each step. This prediction takes the form of a single token, and we maximize the log-likelihood of these target tokens during training. The PRM can therefore be trained in a standard language model pipeline without any special accommodations. To determine the step-level predictions at test time, it suffices to perform a single PRM forward pass over the whole solution. We visualize large-scale PRM scores for two different solutions in \Cref{figure:prm_grading_examples}. To compare multiple solutions, it is necessary to compute a single score for each solution. This is an important but straightforward detail: we define the PRM score for a solution to be the probability that every step is correct under the PRM. We implement this as the product of the correctness probabilities for each step. We describe other possible scoring strategies and additional PRM training details in \Cref{appendix:prm_details}.

When we provide process supervision, we deliberately choose to supervise only up to the first incorrect step. This makes the comparison between outcome and process supervision more straightforward. For correct solutions, both methods provide the same information, namely that every step is correct. For incorrect solutions, both methods reveal the existence of at least one mistake, and process supervision additionally reveals the precise location of that mistake. If we were to provide additional process supervision beyond the first mistake, then process supervision would have an even greater information advantage. This decision also keeps the labelling cost similar for humans: without relying on an easy-to-check final answer, determining the correctness of a solution is equivalent to identifying its first mistake. While most MATH problems do have easy-to-check final answers, we expect this to not remain true in more complex domains.

\section{Large-scale Supervision}
\label{section:large_scale}

We train the large-scale PRM using the step-level labels in PRM800K. To ensure the large-scale ORM baseline is as strong as possible, we train on 100 uniform samples per problem from the generator. This means the ORM training set has no overlap with PRM800K, and it is an order of magnitude larger. Although these two training sets are not directly comparable, each represents our best attempt to advance the state-of-the-art with each form of supervision. We note that training the ORM solely on PRM800K solutions would be problematic, since our active learning strategy has heavily biased the dataset towards wrong-answer solutions. We did explore training the ORM on a superset of PRM800K solutions, by mixing in uniformly sampled solutions, but we found that this did not improve ORM performance.

\begin{figure}
\centering
\begin{subfigure}{\textwidth}
\centering
\begin{tabular}{|{l}|{c}|{c}|{c}|} 
 \hline
  & ORM & PRM & Majority Voting \\ 
 \hline
 \% Solved (Best-of-1860) & $72.4$ & $\mathbf{78.2}$ & $69.6$ \\ 
 \hline
\end{tabular}
\end{subfigure}

\begin{subfigure}{0.825 \textwidth}
\includegraphics[width=\textwidth]{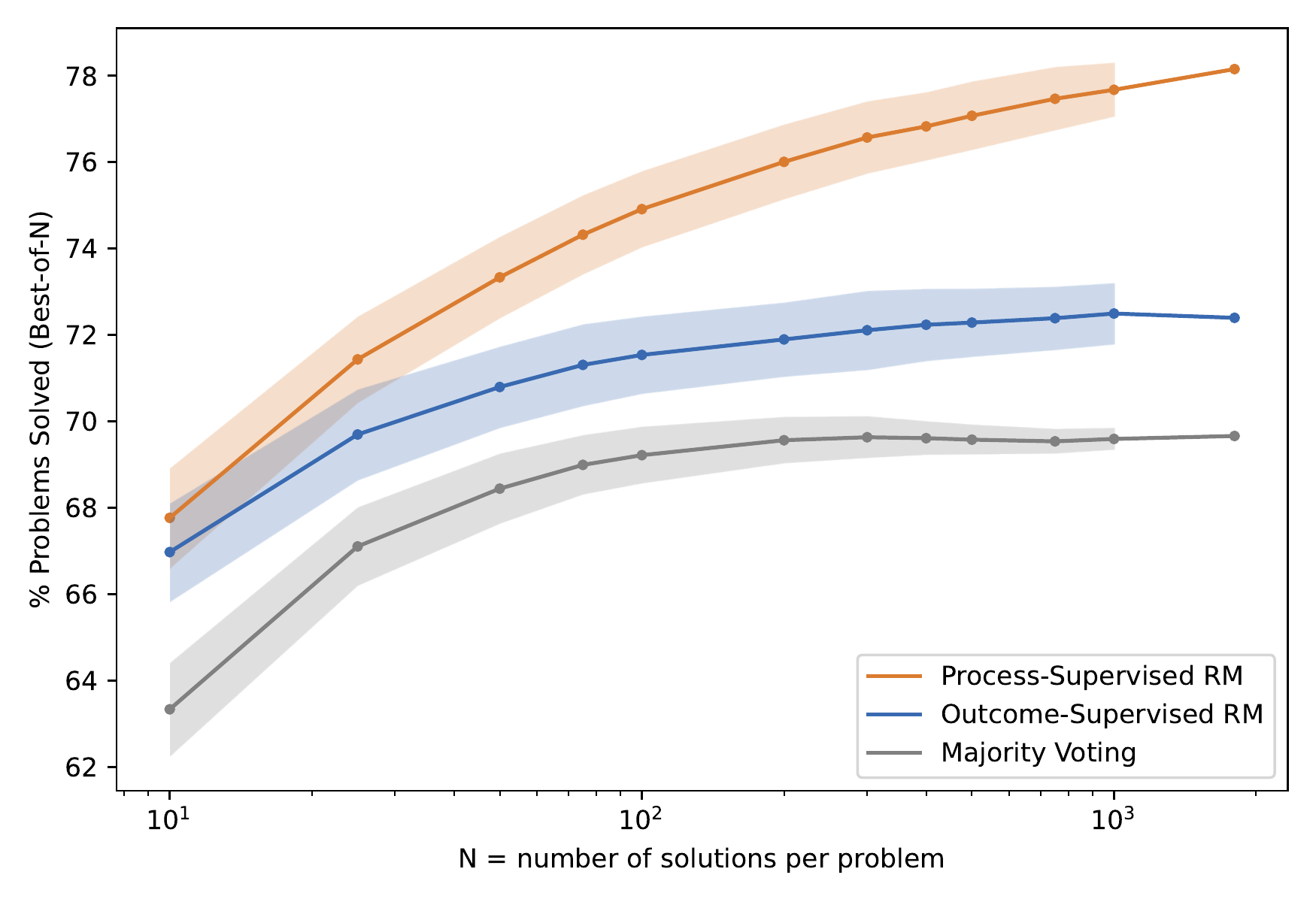}
\end{subfigure}
 
\caption{A comparison of outcome-supervised and process-supervised reward models, evaluated by their ability to search over many test solutions. Majority voting is shown as a strong baseline. For $N \leq 1000$, we visualize the variance across many subsamples of the 1860 solutions we generated in total per problem.}
\label{figure:orm_prm}
\end{figure}

\Cref{figure:orm_prm} shows how the best-of-N performance of each reward model varies as a function of N. Since majority voting is known to be a strong baseline \citep{wang2022self, lewkowycz2022solving}, we also include this method as a point of comparison. While the ORM performs slightly better than the majority voting baseline, the PRM strongly outperforms both. Not only does the PRM reach higher performance for all values of N, but the performance gap widens as N increases. This indicates that the PRM is more effective than both the ORM and majority voting at searching over a large number of model-generated solutions. We experimented with using RM-weighted voting \citep{li2022advance, uesato2022solving} to combine the benefits of the PRM and majority voting, but this did not noticeably improve performance. We use a specific subset of the MATH test set for evaluation, which we describe in \Cref{appendix:evaluation}. We further break down these results by problem difficulty in \Cref{appendix:difficulty_breakdown}.

\section{Small-scale Synthetic Supervision} \label{section:synthetic_supervision}

We find that the PRM outperforms the ORM at large-scale, but this result alone paints an incomplete picture. To better compare outcome and process supervision, there are two confounding factors that must be isolated. First, the training sets for the ORM and the PRM are not directly comparable: the PRM training set was constructed using active learning, is biased towards answer-incorrect solutions, and is an order of magnitude smaller. Second, the final-answer grading will provide positive labels to spurious solutions that reach the correct final answer despite incorrect reasoning. This could damage ORM performance, an effect we may or may not want to attribute to outcome supervision more generally.

Due to the high cost of collecting human feedback, we cannot easily ablate these factors using human labelers. We instead perform the relevant ablations by using the large-scale PRM to supervise smaller models. This setup enables us to simulate a large amount of data collection at a modest cost. For the remainder of this section, we refer to the large-scale PRM from \Cref{section:large_scale} as $\text{PRM}_{\text{large}}$.

\begin{figure}
\centering
\begin{subfigure}{0.45 \textwidth}
\includegraphics[width=\textwidth]{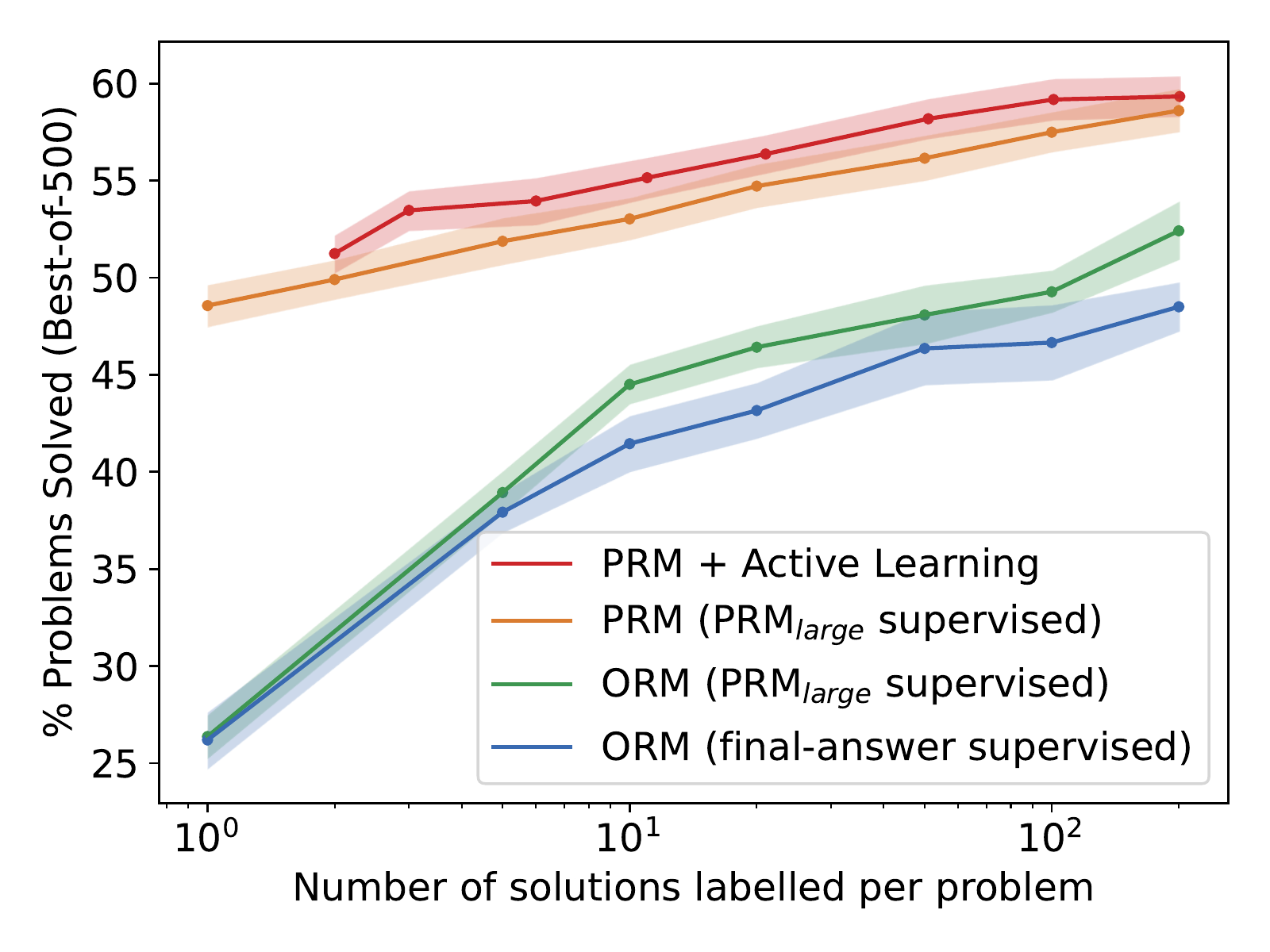}
\caption{Four series of reward models trained using different data collection strategies, compared across training sets of varying sizes.}
\label{figure:small_synthetic_data_scaling}
\end{subfigure}
\hspace{2mm}
\begin{subfigure}{0.45 \textwidth}
\includegraphics[width=\textwidth]{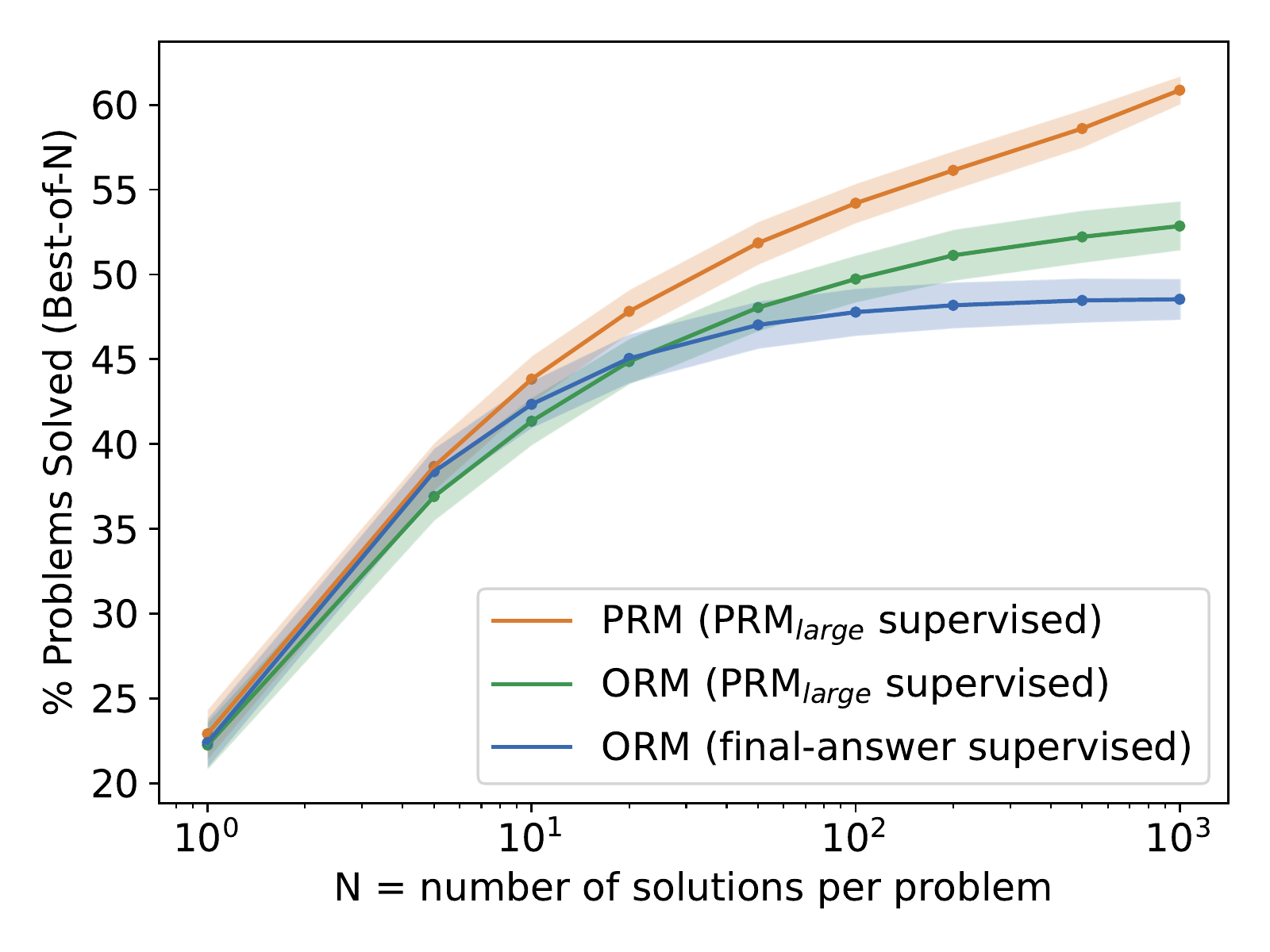}
\caption{Three reward models trained on 200 samples/problem using different forms of supervision, compared across many test-time compute budgets.}
\label{figure:small_synthetic_robustness}
\end{subfigure}
\caption{A comparison of different forms of outcome and process supervision. Mean and standard deviation is shown across three seeds.}
\label{figure:small_synthetic}
\end{figure}

\subsection{Process vs Outcome Supervision}

We now conduct a direct comparison of outcome and process supervision. We first sample between 1 and 200 solutions per problem from a small-scale generator. For each dataset, we provide three forms of supervision: process supervision from $\text{PRM}_{\text{large}}$, outcome supervision from $\text{PRM}_{\text{large}}$, and outcome supervision from final-answer checking. The choice of supervision is the only difference between these three series of reward models, which are otherwise trained on identical datasets. See \Cref{appendix:prm_synthetic} for more details about how $\text{PRM}_{\text{large}}$ is used for outcome and process supervision.

In \Cref{figure:small_synthetic_data_scaling}, we evaluate each reward model by its best-of-500 selection. We see that process supervision significantly outperforms both forms of outcome supervision at all data collection scales. In \Cref{figure:small_synthetic_robustness}, we evaluate the best reward model from each series by its best-of-N performance across different values of N. We see that using $\text{PRM}_{\text{large}}$ for outcome supervision is noticeably more effective than final-answer checking. This can be explained by the fact that $\text{PRM}_{\text{large}}$ provides better supervision for solutions that reach the correct final answer using incorrect reasoning.

It is not clear whether supervision by $\text{PRM}_{\text{large}}$ or by final-answer checking represents the more appropriate outcome supervision baseline. While final-answer supervision is more explicitly outcome based, its main weakness --- the existence of false positives --- is arguably over-emphasized in the MATH dataset. Outcome supervision by $\text{PRM}_{\text{large}}$ better represents outcome supervision in domains that are less susceptible to false positives. We consider outcome supervision by $\text{PRM}_{\text{large}}$ to be the more relevant baseline, but we encourage the reader to draw their own conclusions.

\subsection{Active Learning} \label{section:active_learning}

Finally, we investigate the impact of active learning. We train a small-scale reward model, $\text{PRM}_{\text{selector}}$, on a single sample from each problem, and we use this model to score 1000 samples per problem. To train each of our larger reward models, we select $N$ samples per problem such that $80\%$ are the most convincing (according to $\text{PRM}_{\text{selector}}$) wrong-answer samples, and $20\%$ are the most convincing samples that remain (right- or wrong-answer). We score the selected samples with $\text{PRM}_{\text{large}}$ and train on those scores. This process ensures that all samples are relatively convincing under $\text{PRM}_{\text{selector}}$, that a large fraction are known to contain at least one mistake, and that our overall dataset is not too heavily biased toward wrong-answer solutions. Performance of this data labelling scheme is shown in \Cref{figure:small_synthetic_data_scaling}. By comparing the slopes of the line of best fit with and without active learning, we estimate that this form of active learning is approximately 2.6x more data efficient than uniform data labelling. We note that the model trained on the largest active learning dataset (200 samples per problem) appears to slightly underperform the expected trend line. Our best explanation for this observation is that 200 samples represents a significant fraction of the overall selection pool (1000 samples) and that this relative lack of diversity limits the possible upside from active learning.

We also performed a preliminary investigation into the impact of iteratively retraining $\text{PRM}_{\text{selector}}$ throughout data collection. Between iterations, we re-trained $\text{PRM}_{\text{selector}}$ using all currently labeled data. Unfortunately, we observed instability in this process which we were unable to diagnose. The resulting reward models performed no better than the models described above. We expect some form of iterative retraining to be beneficial in active learning, but we currently have no concrete evidence to support this claim. We consider this a compelling direction for future research.

\section{OOD Generalization} \label{section:generalization}

\begin{table}
\centering
\begin{tabular}{l*{4}{c}} 
 \hline
  & ORM & PRM & Majority Vote & \# Problems \\ 
 \hline
 AP Calculus & $68.9\%$ & $\mathbf{86.7}\boldsymbol{\%}$ & $80.0\%$ & 45 \\ 
 AP Chemistry & $68.9\%$ & $\mathbf{80.0}\boldsymbol{\%}$ & $71.7\%$ & 60 \\
 AP Physics & $77.8\%$ & $\mathbf{86.7}\boldsymbol{\%}$ & $82.2\%$ & 45 \\
 AMC10/12 & $49.1\%$ & $\mathbf{53.2}\boldsymbol{\%}$ & $32.8\%$ & 84 \\
 Aggregate & $63.8\%$ & $\mathbf{72.9}\boldsymbol{\%}$ & $61.3\%$ & 234 \\
 \hline
\end{tabular}
\caption{We measure out-of-distribution generalization using recent STEM tests. We evaluate the outcome-supervised RM, the process-supervised RM, and majority voting using 100 test samples per problem.}
\label{table:generalization}
\end{table}

To get some measure of out-of-distribution generalization, we evaluate our large-scale ORM and PRM on a held-out set of 224 STEM questions, pulled from the most recent AP Physics, AP Calculus, AP Chemistry, AMC10, and AMC12 exams. Since these tests were released after the pre-training dataset was compiled, we can have high confidence that the model has not seen these problems. We report the best-of-100 performance of the ORM, PRM and majority voting in \Cref{table:generalization}. We observe results similar to those in \Cref{section:large_scale}: the PRM outperforms both the ORM and majority voting. This shows us that the PRM can tolerate a modest amount of distribution shift and that its strong performance holds up on fresh test questions.

\section{Discussion} \label{section:discussion}

\subsection{Credit Assignment}

One clear advantage of process supervision is that it provides more precise feedback than outcome supervision. A reward model trained with outcome supervision faces a difficult credit-assignment task --- to generalize well, it must determine where an incorrect solution went wrong. This is particularly difficult for hard problems: most model-generated solutions contain an error somewhere, so the marginal value of a negative label from outcome supervision is low. In contrast, process supervision provides a richer signal: it specifies both how many of the first steps were in fact correct, as well as the precise location of the incorrect step. Process supervision makes credit assignment easier, and we believe that this explains its strong performance.

\subsection{Alignment Impact}

Process supervision has several advantages over outcome supervision related to AI alignment. Process supervision is more likely to produce interpretable reasoning, since it encourages models to follow a process endorsed by humans. Process supervision is also inherently safer: it directly rewards an aligned chain-of-thought rather than relying on outcomes as a proxy for aligned behavior \citep{ought2022}. In contrast, outcome supervision is harder to scrutinize, and the preferences conveyed are less precise. In the worst case, the use of outcomes as an imperfect proxy could lead to models that become misaligned after learning to exploit the reward signal \citep{uesato2022solving, cotra2022, everitt2017reinforcement}. 

In some cases, safer methods for AI systems can lead to reduced performance \citep{ouyang2022training, askell2021general}, a cost which is known as an alignment tax. In general, any alignment tax may hinder the adoption of alignment methods, due to pressure to deploy the most capable model. Our results show that process supervision in fact incurs a negative alignment tax. This could lead to increased adoption of process supervision, which we believe would have positive alignment side-effects. It is unknown how broadly these results will generalize beyond the domain of math, and we consider it important for future work to explore the impact of process supervision in other domains.

\subsection{Test Set Contamination} \label{section:contamination}

The test set of the MATH dataset contains problems that are discussed in several online venues, and it is likely that some of these problems appear in the pretraining dataset for our models. We attempted to remove all MATH problems from our MathMix dataset using string-matching heuristics, but since humans can post hard-to-detect rephrasings of a problem online, it is difficult to make any strong guarantees about the overlap between MathMix and the MATH dataset.

In our experience inspecting model-generated solutions, we saw no clear signs of our models memorizing MATH problems. However, it is impossible to rule out subtle forms of memorization that would slip past manual inspection, and it is still possible that some degree of contamination has slightly inflated our performance on the MATH test set. Even in that case, we would expect any contamination to manifest similarly across all methods, and that the relative comparisons made throughout this work would remain mostly unaffected.

We also note that the PRM regularly surfaces correct solutions to MATH problems that have a low single-digit percentage solve-rate under the generator, some examples of which can be seen in \Cref{appendix:prm_visualizations}. The generator's low solve-rate is an additional indication that it has not encountered such problems via test set contamination. Our generalization results from \Cref{section:generalization} further strengthen our claim that test set contamination has not significantly impacted this work, since we observe qualitatively similar results on problems that are guaranteed to be uncontaminated.

\section{Related Work}

\subsection{Outcome vs Process Supervision}

In work closely related to our own, \cite{uesato2022solving} compare the impact of outcome and process supervision in the domain of grade school math. They found that both methods led to similar final-answer error rates, and that process supervision achieved those results with less data. While our core methodology is very similar, there are three main details that differ. First, we use a more capable model to collect PRM800K dataset and to perform our large-scale experiments. However, our small-scale results in \Cref{section:synthetic_supervision} suggest that large-scale models are not necessary to observe benefits from process supervision. Second, we evaluate on the MATH dataset, which is significantly more challenging than GSM8K. Third, we collect a much larger quantity of process supervision data.

On the surface, the results from \cite{uesato2022solving} may seem to conflict with our claim that process supervision leads to better performance. However, we believe the apparent conflict can be explained by the difference in the scale of the supervision. The data scaling trend in \Cref{figure:small_synthetic_data_scaling} suggests that a small amount of process supervision and a large amount of outcome supervision do in fact lead to similar performance, consistent with the results from \cite{uesato2022solving}. The trend also shows that process supervision beats outcome supervision when scaled up, even when judged based solely on outcomes. This is consistent with our results in \Cref{section:large_scale}. We believe these results make a strong case for using process supervision.

\subsection{Synthetic Supervision}

Similar to our work in \Cref{section:synthetic_supervision}, \cite{gao2022scaling} use a large reward model to supervise the training of smaller models. They study the over-optimization that occurs during RLHF, with experiments that require large quantities of human preference data. To work around this challenge, they use a gold-standard reward model to replace human feedback. Our use of a large-scale reward model to supervise smaller reward models shares similarities with their approach.

\subsection{Natural Language Reasoning}

Several recent studies that have examined the reasoning ability of large language models are implicitly relevant to our work. \cite{lewkowycz2022solving} showed that finetuning models on a large corpus of technical content led to significantly improved performance on MATH. \cite{wang2022self} showed that \textit{self-consistency} leads to remarkably strong performance on many reasoning benchmarks, notably without requiring any additional finetuning. \cite{wei2022chain} and \cite{nye2021show} demonstrate the importance of explicitly performing intermediate reasoning steps via a \textit{chain of thought} or a \textit{scratchpad} in order to solve tasks that require multi-step reasoning. \cite{kojima2022large} show that models are able to perform this behavior zero-shot, conditioned only on a simple prompt.

\section{Conclusion}

We have shown that process supervision can be used to train much more reliable reward models than outcome supervision in the domain of mathematical reasoning. We have also shown that active learning can be used to lower the cost of human data collection by surfacing only the most valuable model completions for human feedback. We release PRM800K, the full dataset of human feedback used to train our state-of-the-art reward model, with the hope that removing this significant barrier to entry will catalyze related research on the alignment of large language models. We believe that process supervision is currently under-explored, and we are excited for future work to more deeply investigate the extent to which these methods generalize.

\section*{Acknowledgements}

We thank Joshua Achiam, Mark Chen, Jonathan Gordon, Dan Hendrycks, Lukasz Kaiser, Oleg Murk, Ben Sokolowsky, Francis Song, and Jonathan Uesato for valuable feedback and thoughtful discussions; Giambattista Parascandolo and Daniel Selsam for their contributions to the MathMix dataset; Jonathan Ward for contributing to the data collection interface; Wojciech Zaremba for encouraging us to scale up data collection; Peter Hoeschele and Aris Kostantinidis for supporting our data collection; the research acceleration and supercomputing teams at OpenAI for providing infrastructure support; and the team at Scale and the many data-labelers who created PRM800K.

\changeurlcolor{black}

\bibliography{process_supervision}

\changeurlcolor{blue}

\appendix

\clearpage

\section{MathMix} \label{appendix:mathmix}

Similar to \cite{lewkowycz2022solving} we construct a large-scale dataset of high-quality math-relevant tokens for use in a lightweight pretraining stage, before finetuning on comparably smaller datasets like MATH and PRM800K. This dataset, which we call MathMix, has two main differences compared to the one used to train Minerva. First, it is smaller and more aggressively filtered to high-quality math problem-solving content, and second, it does not explicitly mix in general language data.

Minerva was trained on 38.5B tokens of arXiv documents and webscrape pages with LaTeX content, while MathMix consists of a smaller set of 1.5B tokens containing individual math problems and their solutions, free-form text discussing math problems and concepts, and synthetic data (Table \ref{table:mathmix}). While Minerva was pretrained on a dataset with 5\% general natural language data, we chose not to mix in any natural language data explicitly, primarily because MathMix already contains plenty of natural language data.

\begin{table}[!h]
\centering
\begin{tabular}{l*{2}{c}} 
 \hline
  Data type & Token count & Present in pretraining? \\ 
 \hline
 Math problems and solutions & $\sim$ 275M & No \\ 
 Free-form math discussion text (1) & $\sim$ 430M & No \\
 Free-form math discussion text (2) & $\sim$ 450M & Yes \\
 Synthetic data (1) & $\sim$ 30M & No \\
 Synthetic data (2) & $\sim$ 100M & Yes \\
 Critiques grading data & $\sim$ 500M & No \\
 \hline
\end{tabular}
\caption{MathMix dataset components.}
\label{table:mathmix}
\end{table}

Note that when training smaller models, as in Section \ref{section:synthetic_supervision}, we use a slightly smaller variant of MathMix that excludes the critiques data and only consists of 1B tokens. For our large models experiments, we train on MathMix for roughly 3B tokens (2 epochs). For our small models experiments, we train for 6 epochs (roughly 6.6B tokens).

We apply a set of decontamination checks on MathMix against the test split of the MATH dataset, including stripping out LaTeX and searching for matching n-grams, but we can make no strong guarantees on the efficacy of this decontamination. As discussed in \Cref{section:contamination}, we would not expect the relative comparisons made throughout this work to be significantly impacted by test set contamination.

\newpage

\section{PRM800K} \label{appendix:data_collection}

We collected $1,\!085,\!590$ step-level labels over $101,\!599$ solution samples. We present the whole unfiltered dataset as PRM800K. During training we discard labels used for quality control, as well as any step-level labels for which the labeler was unable to complete the task. The filtered dataset contains about $800,\!000$ step-level labels over $75,\!000$ solutions. The full PRM800K dataset is available at \href{https://github.com/openai/prm800k}{https://github.com/openai/prm800k}.

The data collection was split into two separate phases. In phase 1, we collected labels for multiple alternative completions at each step of a solution. This seeded our dataset but was cumbersome---for many steps the alternatives were repetitive, and we found labelers spent a lot of time supervising long uninteresting solutions. As a result, the step-level labels we collected in this phase are more repetitive than those collected later. In total, phase 1 represents about $5\%$ of PRM800K, or about $40,\!000$ step-level labels.

The majority of our labels were collected as part of phase 2, during which we scaled up and streamlined the data collection process. Phase 2 data collection is split into $10$ generations. For each generation, we sample $N$ solutions per problem from the generator. We rank these solutions with our current best PRM and surface the highest scoring wrong-answer solutions to our labelers. We retrain this PRM between each generation using all the latest data. This active learning strategy changes the balance of our data considerably. Though we sometimes surfaced correct solutions (either by manually injecting correct solutions or because of errors in our automatic grading), the vast majority of the labels we collected in this phase are for incorrect solutions. \Cref{table:balance} breaks down the balance of correct/incorrect steps and solutions between the different phases of data collection. Though we mostly collected labels on incorrect solutions, we still collected many labels for correct individual steps. In fact, our small-scale ablations in \Cref{section:active_learning} suggest that this active learning strategy, which favors labelling high-scoring wrong-answer solutions, improves performance despite the resulting imbalance in the dataset.

\begin{table}[!h]
\centering
\begin{tabular}{l*{3}{c}} 
 \hline
  & phase $1$ & phase $2$ & combined \\ 
 \hline
 $\%$ end in correct solution & $85.1$ & $13.2$ & $14.2$ \\ 
 $\%$ correct steps & $58.6$ & $74.1$ & $73.1$ \\ 
 \hline
\end{tabular}
\caption{Distribution of positive/negative steps/solutions.}
\label{table:balance}
\end{table}

Some of our phase 2 questions are intended for quality control. For a quality control question, researchers mark which steps are reasonable to label as incorrect. Then we assess that labelers are able to consistently mark those steps as incorrect. Prior to starting on phase 2, we required all labelers to label 30 quality control questions. This served as a screening test, and we only admitted labelers that agreed with our gold labels at least $75\%$ of the time.

We then designated $10$-$20$ problems per generation as additional quality control questions, and we randomly served them to labelers as they worked through the task. We used the results of this continuous quality control to remove labelers whose quality slipped too far, as well as to prepare educational material on common mistakes in order to improve labeler alignment with our instructions.

\section{Evaluation} \label{appendix:evaluation}

\begin{figure}[!h]
\centering
\begin{subfigure}{0.475 \textwidth}
\includegraphics[width=\textwidth]{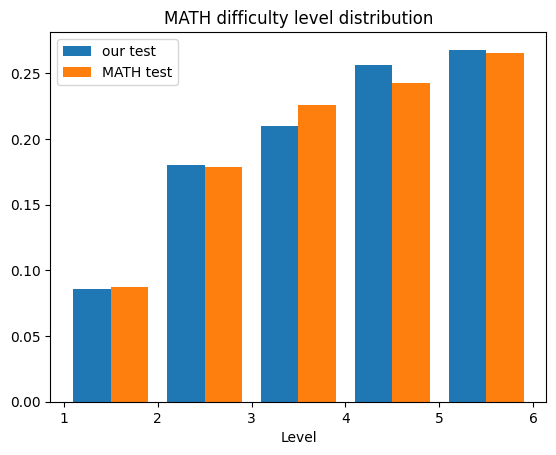}
\vspace{.7cm}
\end{subfigure}
\hspace*{\fill}
\begin{subfigure}{0.475 \textwidth}
\includegraphics[width=\textwidth]{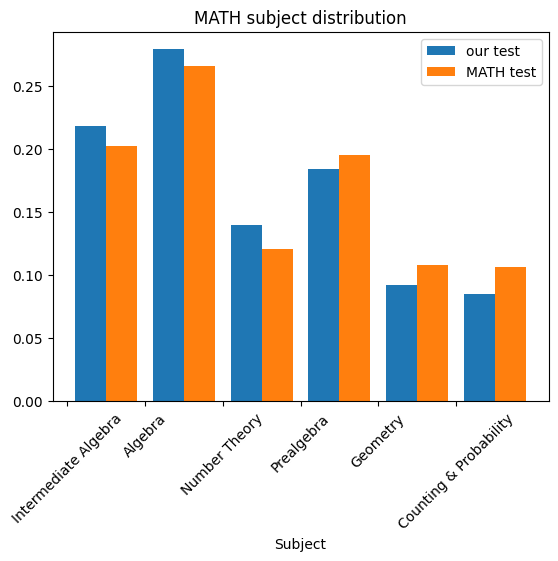}
\end{subfigure}
\caption{Two histograms comparing the distribution of problem difficulty levels and subjects in both the original MATH test set and in our 500 problem test subset.}
\label{figure:math_histogram}
\end{figure}

As we scaled up the project, we began having to collect labels on multiple solutions for the same training problem. In order to avoid the risk of over-fitting on the $7,\!500$ MATH training problems, we expanded the training set to include $4,\!500$ MATH test split problems. We therefore evaluate our models only on the remaining $500$ held-out problems. We selected these $500$ test problems uniformly at random. In \Cref{figure:math_histogram}, we show that the distribution of difficulty levels and subjects in this subset is representative of the MATH test set as a whole. The specific test set we used can be found at \href{https://github.com/openai/prm800k}{https://github.com/openai/prm800k}. We leave it for future work to explore how many distinct training problems are actually necessary, and how quickly our methods overfit to the training set.

\newpage

\section{Labelling Instructions} \label{appendix:labelling_instructions}

Labelers were tasked to look at steps in a solution and label each one as \textbf{positive}, \textbf{negative}, or \textbf{neutral}. A step is considered \textbf{neutral} if it is appropriate in context, reasonable, correct, and contains only computations that can be verified easily. A step is \textbf{positive} if it is \textbf{neutral} and also progresses towards the solution. All other steps are considered \textbf{negative}. Labelers were not given reference solutions, but they were given the ground truth final answers. We chose not to provide reference solutions to avoid biasing them towards one particular path to the solution. We chose to provide ground truth final answers since this information can sometimes help labelers resolve their own misunderstandings.

In phase 1, labelers were permitted to enter their own steps in the case that all candidate steps were \textbf{negative}. Then the solution would progress from a randomly selected \textbf{positive} step (or \textbf{neutral} if their were no \textbf{positive} ones). This often resulted in trajectories that got stuck in endless sequences of \textbf{neutral} steps that said reasonable things but made frustratingly slow progress towards a solution or \textbf{negative} steps that needed constant human supervision. In phase 2, we pre-generate whole solutions and end the task as soon as the first \textbf{negative} step is encountered. The full instructions given to labelers can be found at \href{https://github.com/openai/prm800k/tree/main/prm800k/instructions}{https://github.com/openai/prm800k/tree/main/prm800k/instructions}.

\section{ORM Training Details} \label{appendix:orm_details}

We train outcome-supervised reward models in the same manner as token-level verifiers from \cite{cobbe2021training}, with a few subtle differences to hyperparameters. In particular, we only train for a single epoch on each dataset of model samples and reward model labels, without dropout, and without jointly learning a language modeling objective. We find that performance is not sensitive to most other hyperparameters, within a reasonable range.

To collect model samples, we simply sample uniformly from the generator at a temperature of 1.0 without applying any rebalancing of positives or negatives. At training time, the reward model makes predictions for every token in the context. The target for each token in a solution is the same, based on whether the solution is labelled correct or incorrect. At test time, we simply use the score of the final token in the completion as the overall score of the solution. We note that this setup is identical to the way token-level verifiers were trained in \cite{cobbe2021training}.

\newpage

\section{PRM Details} \label{appendix:prm_details}

\subsection{Training}

We train our PRMs by fine-tuning the MathMix model to predict the probability of positive, negative, and neutral labels given a solution prefix ending in one of our labeled steps. We sweep over hyperparameters using a dataset containing the first $\sim\!10\%$ of PRM800K. Fine-tuning an LLM from its ordinary language modeling task to a classification task like this is a large distribution shift, and we found low learning rates were important to stable PRM training.

All of our PRMs are trained for 2 epochs. On smaller datasets (such as in phase 1 and the first few generations of phase 2) this improves the final performance over training for just 1 epoch. Additional epochs, up to some point, don't noticeably help or hurt performance. On larger datasets, the benefits of 2 epoch training diminishes, but we continue doing it for consistency.

\subsection{Scoring}

There are multiple ways of using the PRM to score solutions. In general, we produce a single solution-level score by performing a reduction over step-level scores, where the step-level score is the probability that the step's label is positive. This involves two specific implementation decisions. First, when determining a step-level score, we either consider a neutral label to be positive or negative. Second, when determining a solution-level score, we either use the minimum or the product over step-level scores as a reduction.

We show results from all four scoring strategies in \Cref{table:scoring}. The best performing strategy is to take the product of step-level scores and to consider the neutrals as positives, but the difference in performance between all strategies is minor. Throughout the rest of this work, we consider neutral steps to be positive, and we define the solution score to be the product of step-level scores. Using the product instead of the minimum as the reduction does create a slight bias against solutions with a larger number of steps.

\begin{table}[!h]
\centering
\begin{tabular}{l*{2}{c}} 
 \hline
  & product & minimum \\ 
 \hline
 neutral = positive & $78.2\%$ & $77.6\%$ \\ 
 neutral = negative & $77.4\%$ & $77.8\%$ \\ 
 \hline
\end{tabular}
\caption{Best-of-1860 test performance using the PRM with four different scoring strategies.}
\label{table:scoring}
\end{table}

\newpage

\section{Difficulty Breakdown} \label{appendix:difficulty_breakdown}

We show performance of our ORM and PRM on each quintile of the MATH dataset. We determine quintiles based on the pass rate under the generator. It is interesting to note that the performance gap is not only apparent on high difficulty problems: it is in fact apparent across all difficulties. For the lowest difficulty problems, we see that it is possible to find adversarial examples that fool the ORM, since the ORM's performance slightly decreases as the number of samples increases. In contrast, the PRM remains highly robust over this same set of samples.

We also see that increasing the number of samples has the largest positive effect on the highest difficulty problems. This is to be expected, since a large number of generator samples may be required to find a true and convincing solution to a hard problem.

\begin{figure}[!h]
\centering
\begin{subfigure}{\textwidth}
\includegraphics[width=\textwidth]{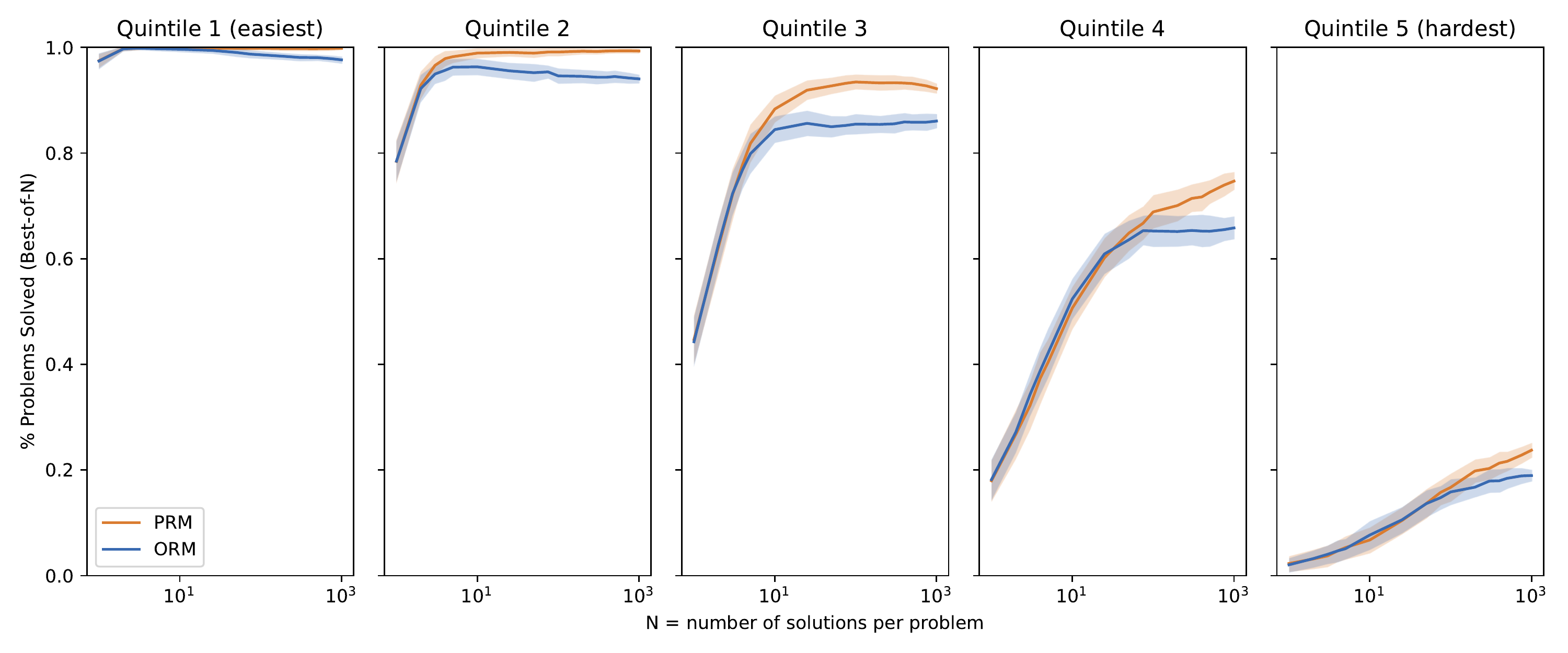}
\end{subfigure}
\caption{A breakdown of ORM vs PRM performance by problem difficulty.}
\label{figure:difficulty}
\end{figure}

\newpage

\section{Synthetic Supervision Details} \label{appendix:prm_synthetic}

We can use $\text{PRM}_{\text{large}}$ to provide either outcome or process supervision for smaller models. We determine the labels for individual steps based on the step-level probabilities outputted by $\text{PRM}_{\text{large}}$. To do this, we set an arbitrary threshold: any step that $\text{PRM}_{\text{large}}$ assigns a negative label with greater than 20\% probability is considered incorrect. We choose this threshold based on the observation that $\text{PRM}_{\text{large}}$ is slightly miscalibrated in the direction of favoring positive labels.

To provide process supervision for a solution, we directly return the step-level labels (positive or negative) provided by $\text{PRM}_{\text{large}}$, up until the first step that is marked as negative. This mimics our true human data collection process. To provide outcome supervision, we mark the solution as correct if and only if $\text{PRM}_{\text{large}}$ considers every step to be correct (using the same thresholding logic).

\newpage

\section{PRM Visualizations} \label{appendix:prm_visualizations}

All examples shown come from the large-scale generator (GPT-4). We note the pass-rate under the generator to give some sense of the difficulty of these problems.

\subsection{True Positives}

These cherry-picked examples show the best-of-1860 solution from the generator as ranked by the large-scale PRM.

\vspace{.5cm}

\noindent
Problem 1. Generator pass-rate: $0.1\%$. This challenging trigonometry problem requires applying several identities in a not-at-all obvious succession. Most solution attempts fail, because it is hard to choose which identities are actually helpful. Though successful solutions to this problem are rare, the reward model correctly recognizes when a valid chain-of-thought has been found.

\vspace{.25cm}

\begin{figure}[!h]
\centering
\includegraphics[width=.8 \textwidth]{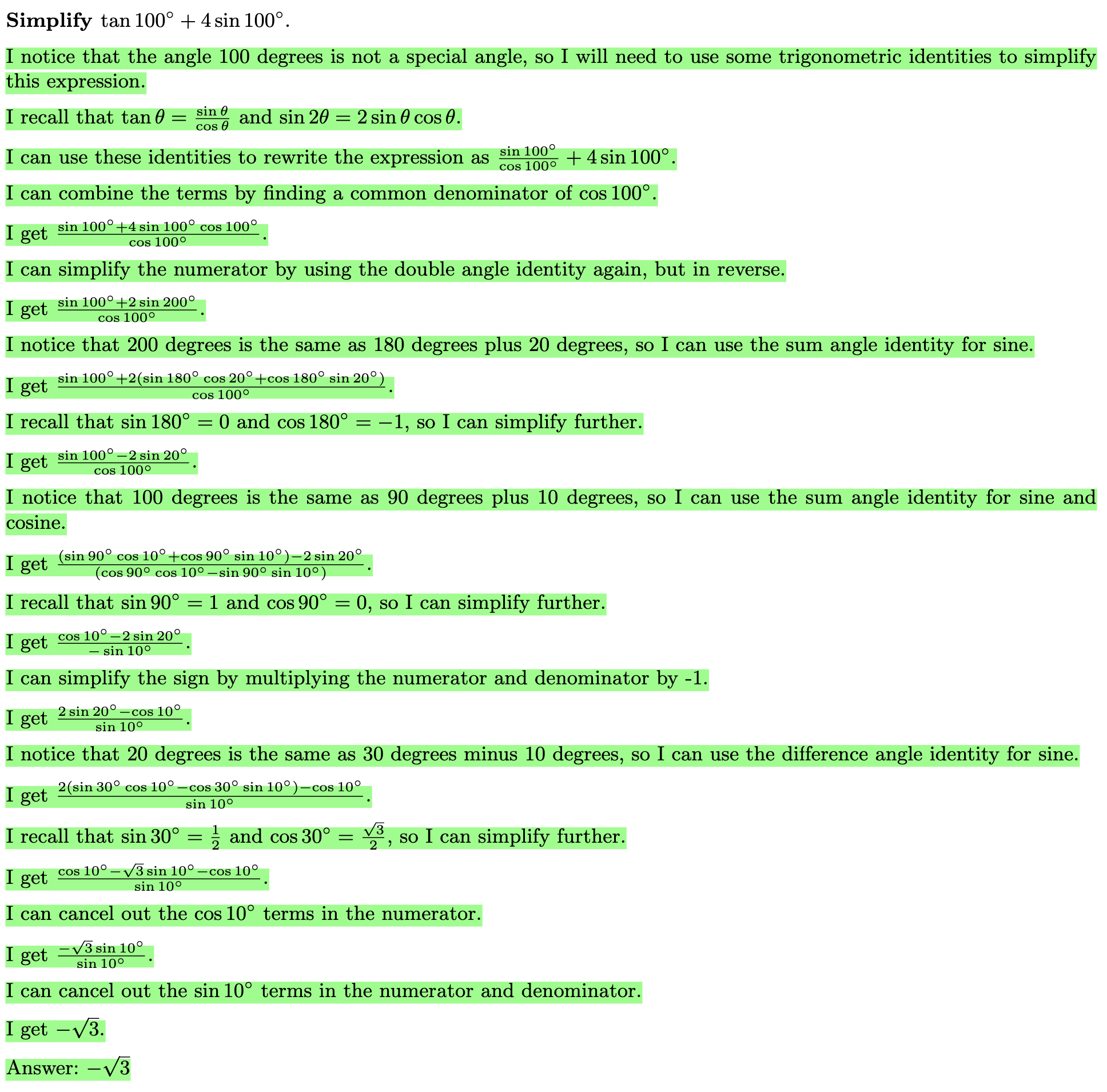}
\end{figure}

\newpage

\noindent
Problem 2. Generator pass-rate: $5.8\%$. In step 7 and 8, the generator starts performing guess-and-check. This is a common place the model might hallucinate, by claiming a particular guess is successful when it isn't. In this case, the reward model verifies each step and determines that the chain-of-thought is correct.

\vspace{.25cm}

\begin{figure}[!h]
\centering
\includegraphics[width=.8 \textwidth]{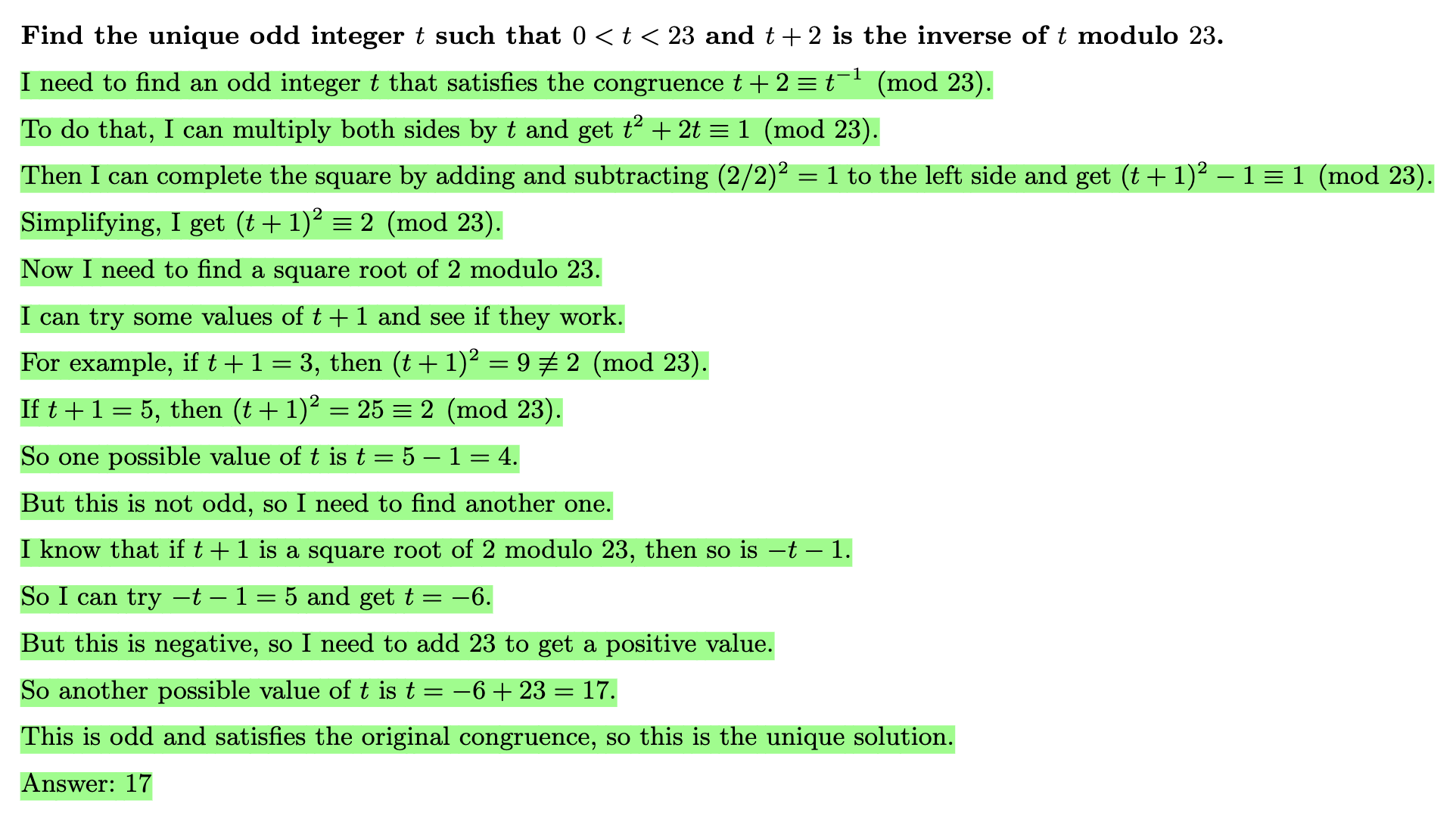}
\end{figure}

\vspace{1cm}

\noindent
Problem 3. Generator pass-rate: $1.7\%$. The generator successfully applies several trigonometric identities to simplify the expression.

\vspace{.25cm}

\begin{figure}[!h]
\centering
\includegraphics[width=.8 \textwidth]{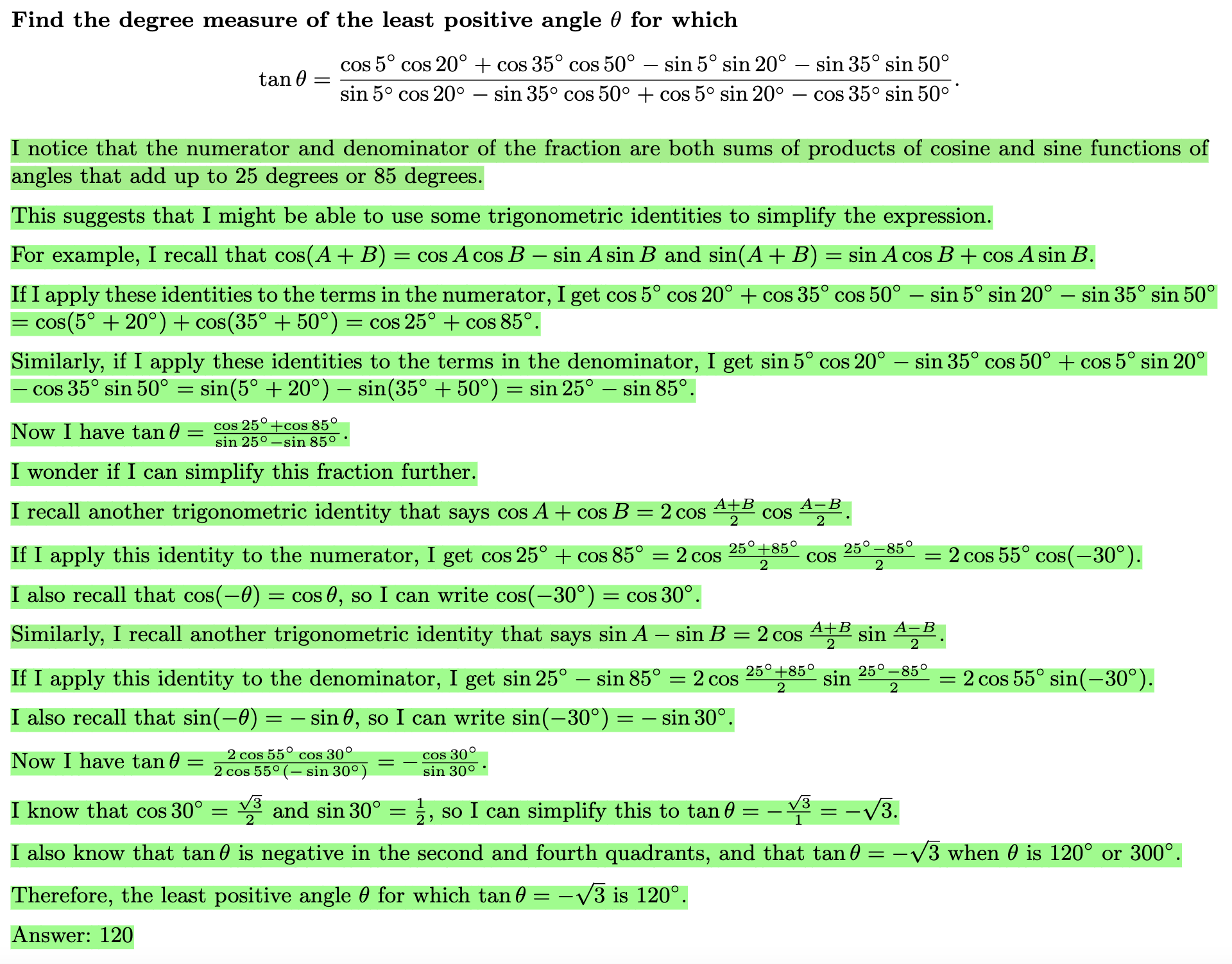}
\end{figure}

\newpage

\noindent
Problem 4. Generator pass-rate: $4.5\%$. Here, the generator successfully performs a complex series of polynomial factorizations. The use of the Sophie-Germain identity in step 5 is an important step that could be considered insightful.

\vspace{.25cm}

\begin{figure}[!h]
\centering
\includegraphics[width=.7 \textwidth]{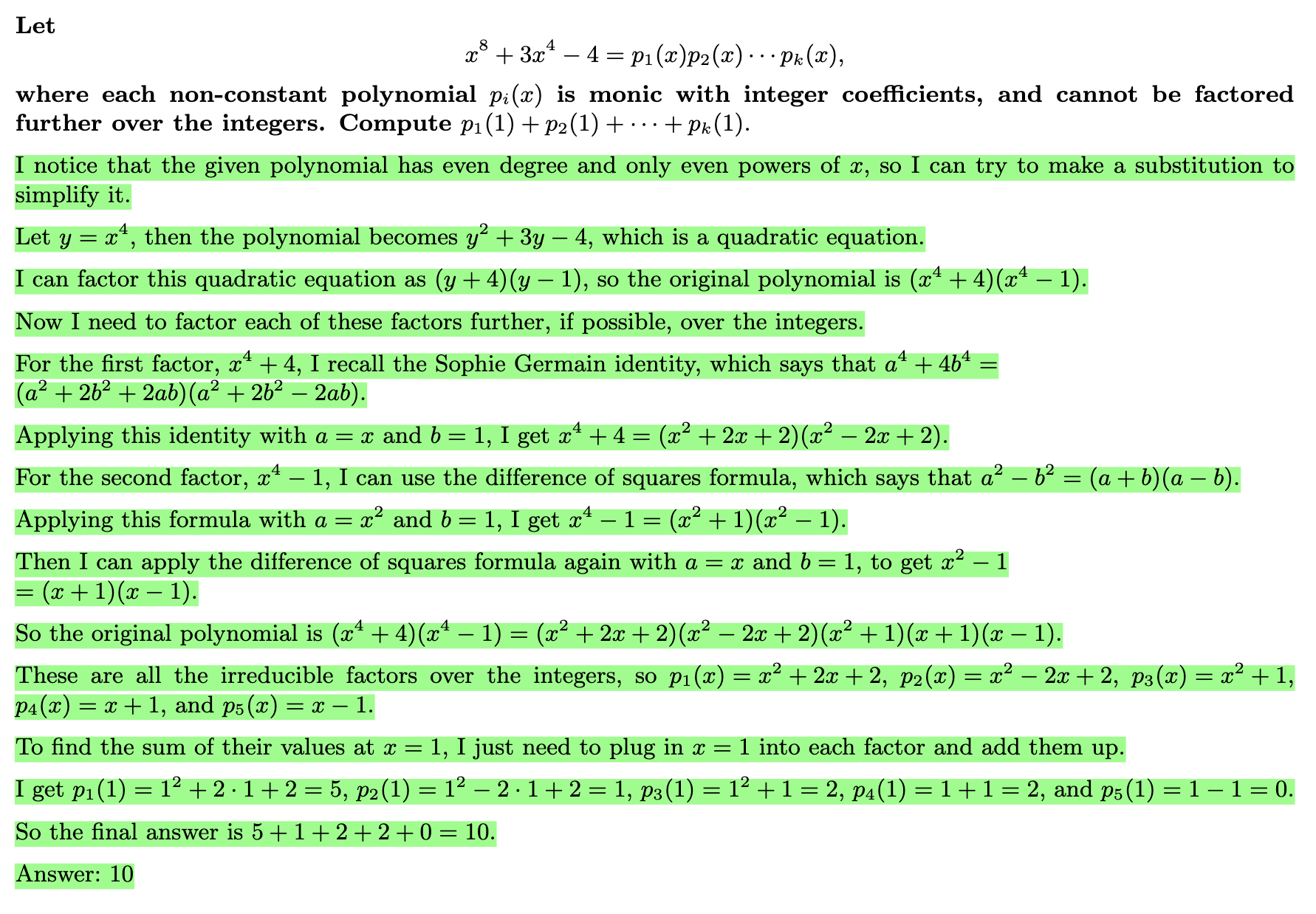}
\end{figure}

\subsection{True Negatives}

\noindent
Problem 5. Generator pass-rate: $4.5\%$. The generator attempts to use the difference of squares formula in step 12 on an expression that isn’t in fact a difference of squares. The reward model catches this mistake.

\vspace{.25cm}

\begin{figure}[!h]
\centering
\includegraphics[width=.7 \textwidth]{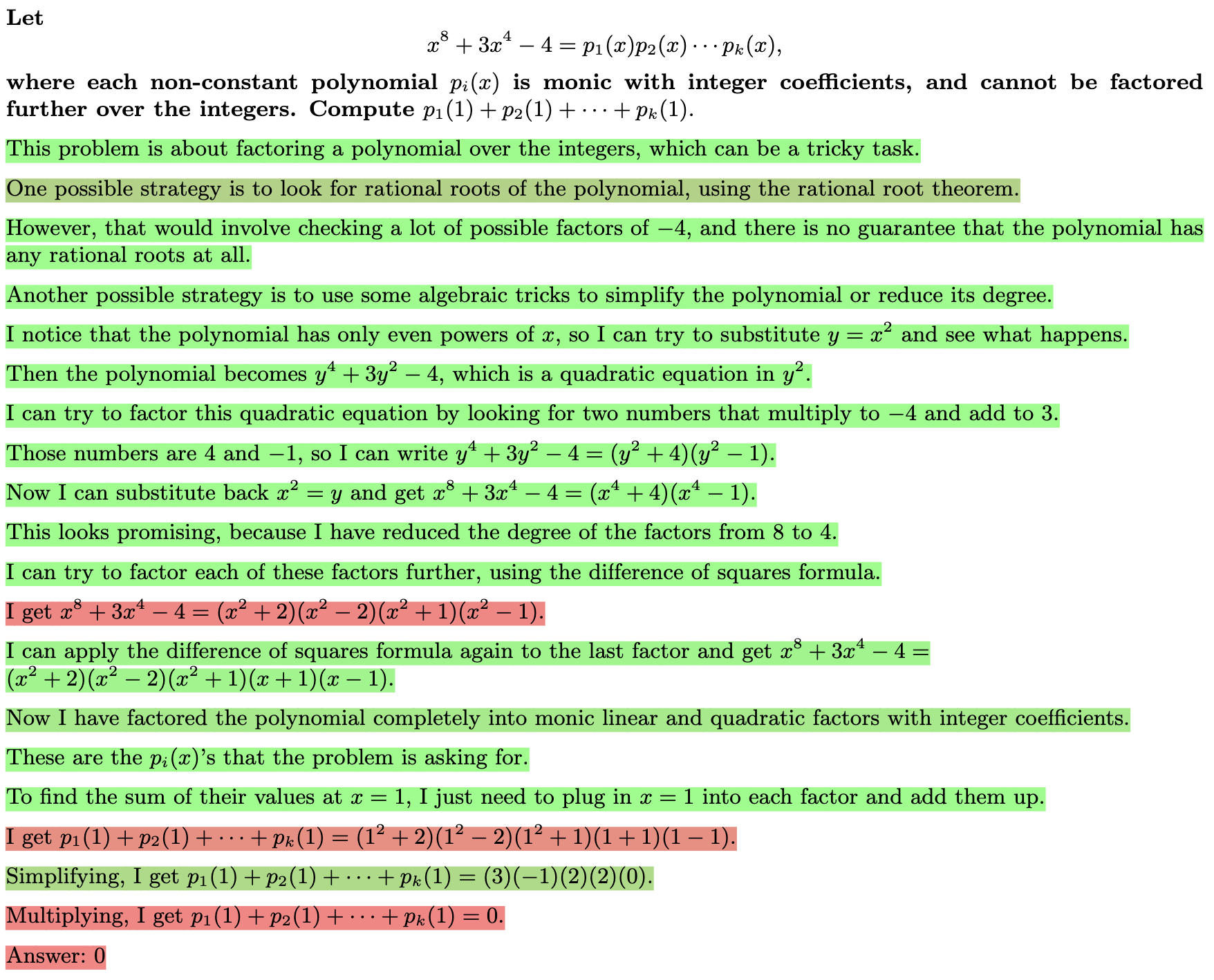}
\end{figure}

\newpage

\noindent
Problem 6. Generator pass-rate: $93.5\%$. In step 7, the generator makes an incorrect attempt to simplify an expression. The reward model catches this mistake.

\vspace{.25cm}

\begin{figure}[!h]
\centering
\includegraphics[width=.8 \textwidth]{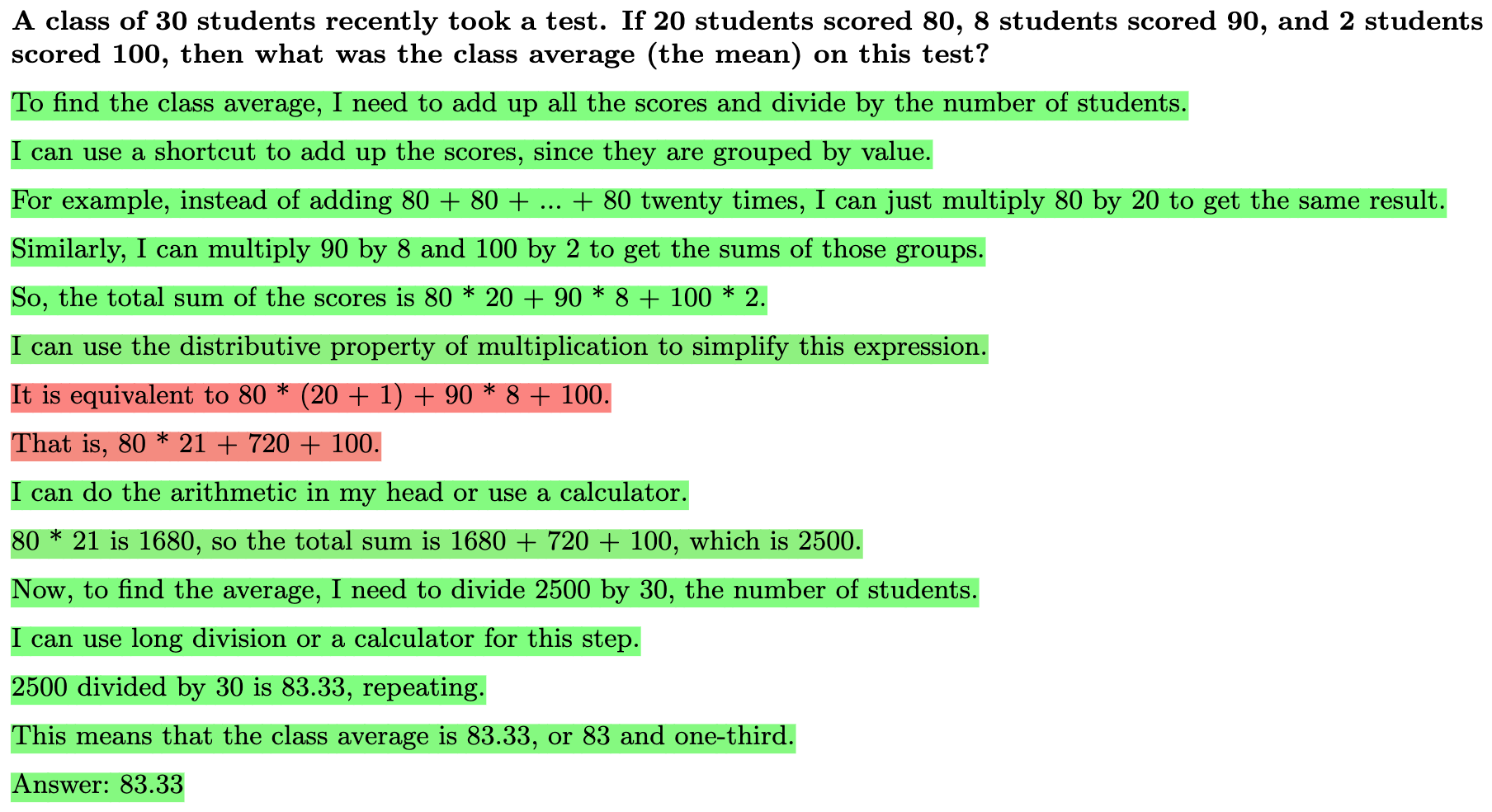}
\end{figure}

\vspace{1cm}

\noindent
Problem 7. Generator pass-rate: $48.0\%$. In step 11, the generator makes a simple calculation error. The reward model catches this mistake.

\vspace{.25cm}

\begin{figure}[!h]
\centering
\includegraphics[width=.8 \textwidth]{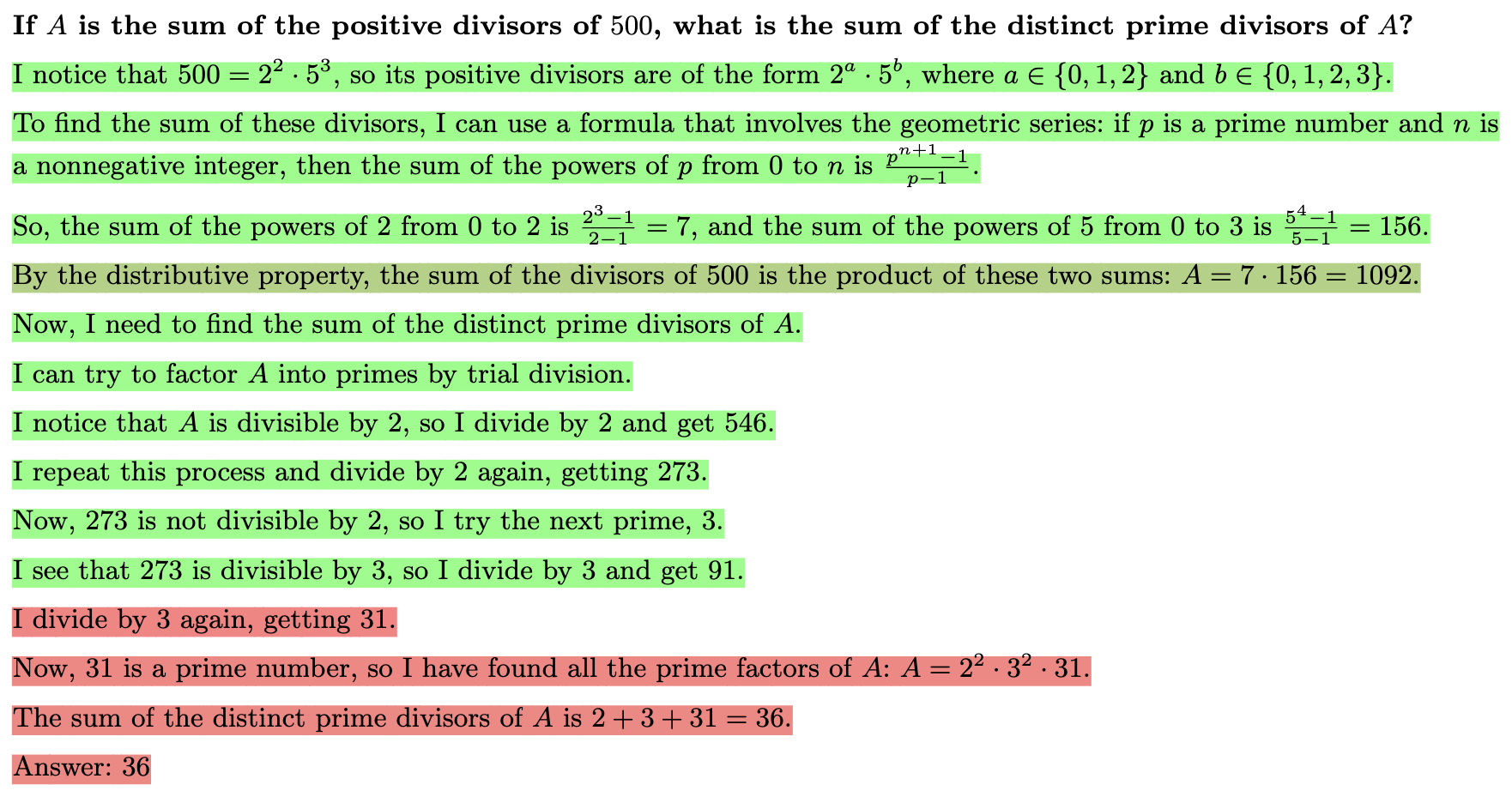}
\end{figure}

\newpage

\noindent
Problem 8. Generator pass-rate: $5.8\%$. The justification in step 8 is strange, but the reward model lets it slide. In step 9, though, the model incorrectly factors the expression. The reward model catches this mistake.

\vspace{.25cm}

\begin{figure}[!h]
\centering
\includegraphics[width=.8 \textwidth]{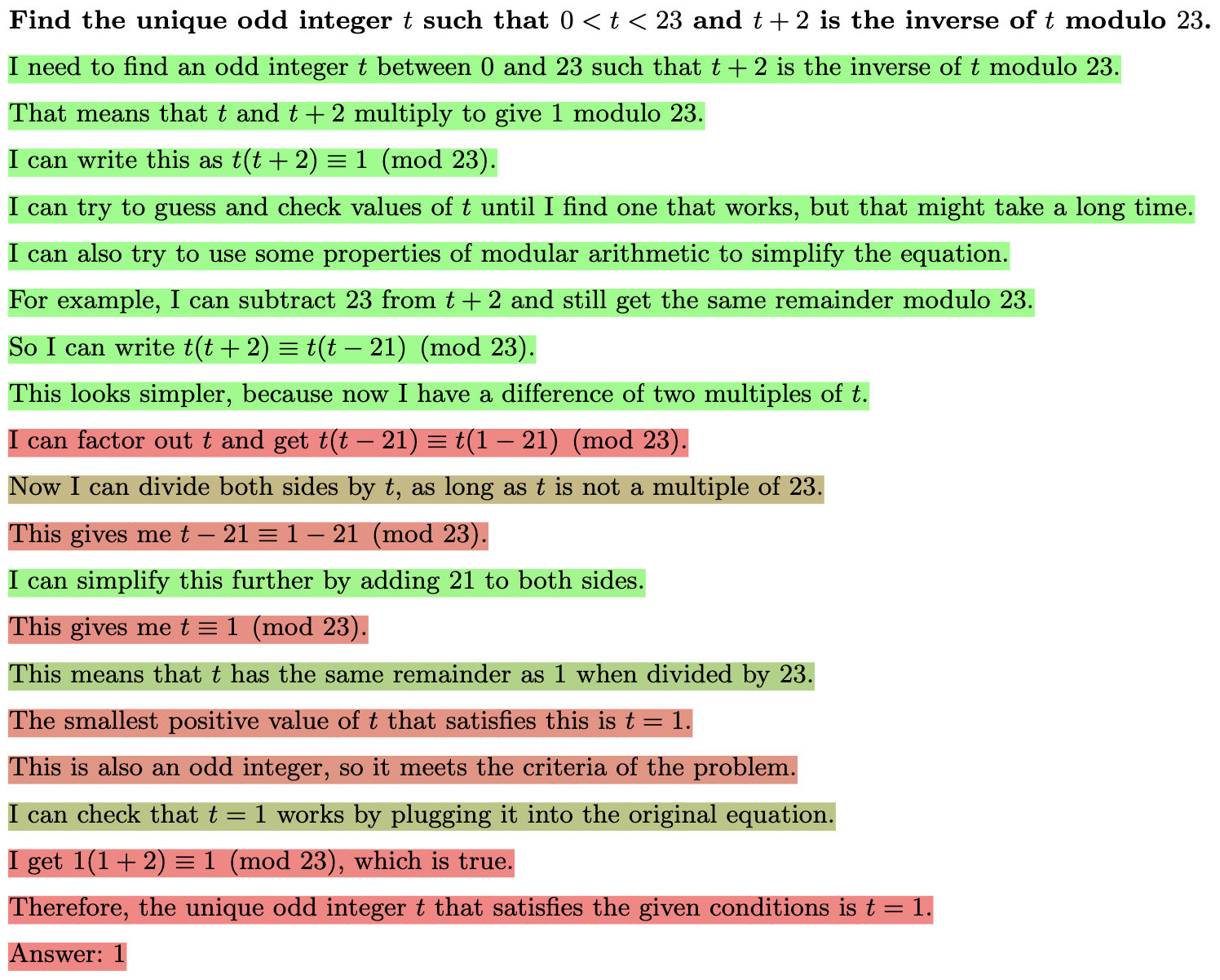}
\end{figure}

\subsection{False Positives}

Problem 9. Generator pass-rate: $18.5\%$. The generator makes a subtle counting error in step 9. On the surface, it appears reasonable to claim that there are 5 ways to exchange the same colored ball since there are 5 colors. However, this undercounts by a factor of 2, since Bob has 2 choices for which ball to return to Alice. The reward model is fooled by this mistake.

\vspace{.25cm}

\begin{figure}[!h]
\centering
\includegraphics[width=.8 \textwidth]{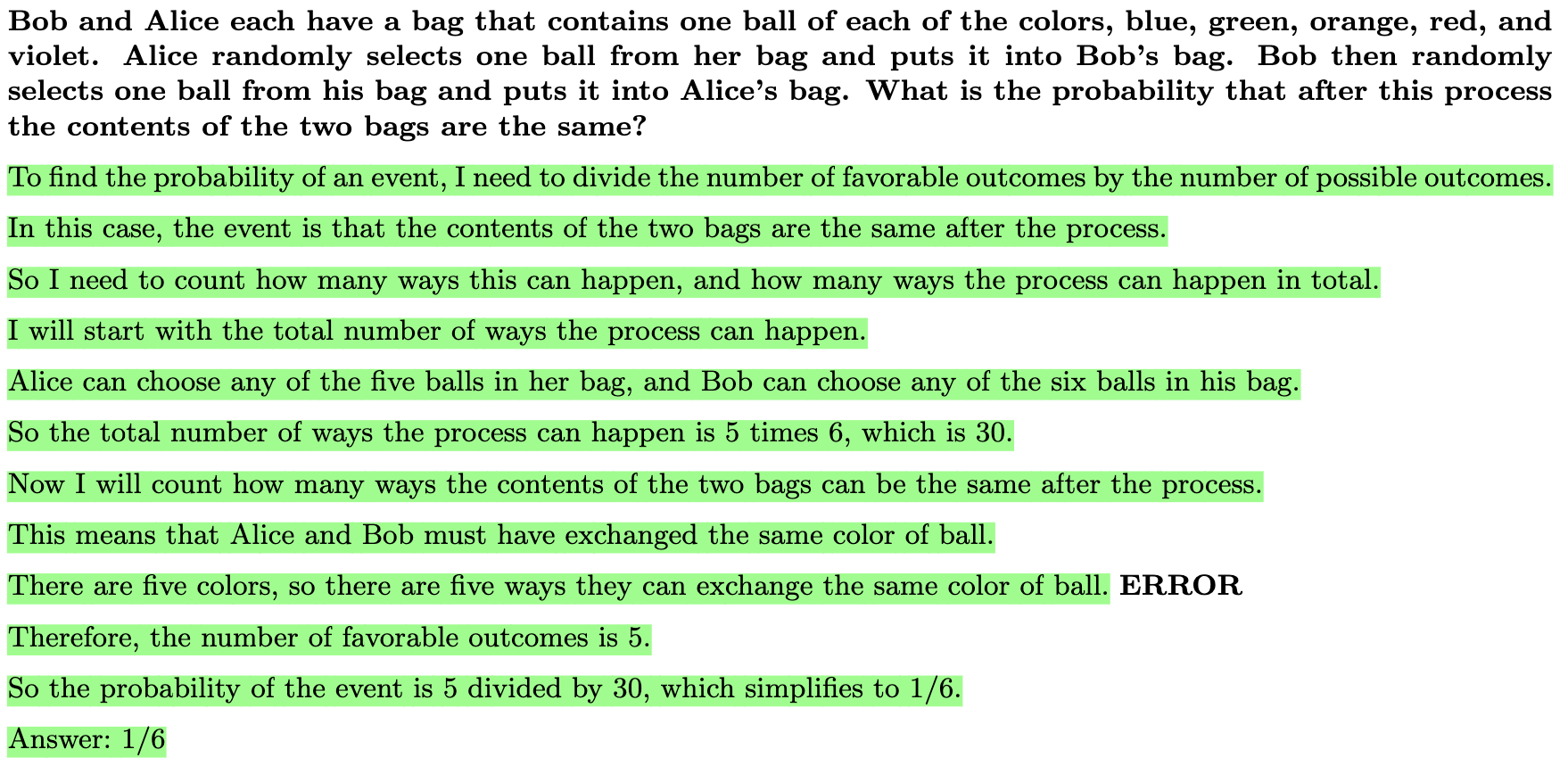}
\end{figure}

\newpage

Problem 10. Generator pass-rate: $17.6\%$. In step 13, the generator attempts to simplify the equation by combining like terms. It correctly moves and combines the linear terms to the left-hand side, but then mistakenly leaves the right-hand side untouched. The reward model is fooled by this mistake.

\vspace{.25cm}

\begin{figure}[!h]
\centering
\includegraphics[width=.7 \textwidth]{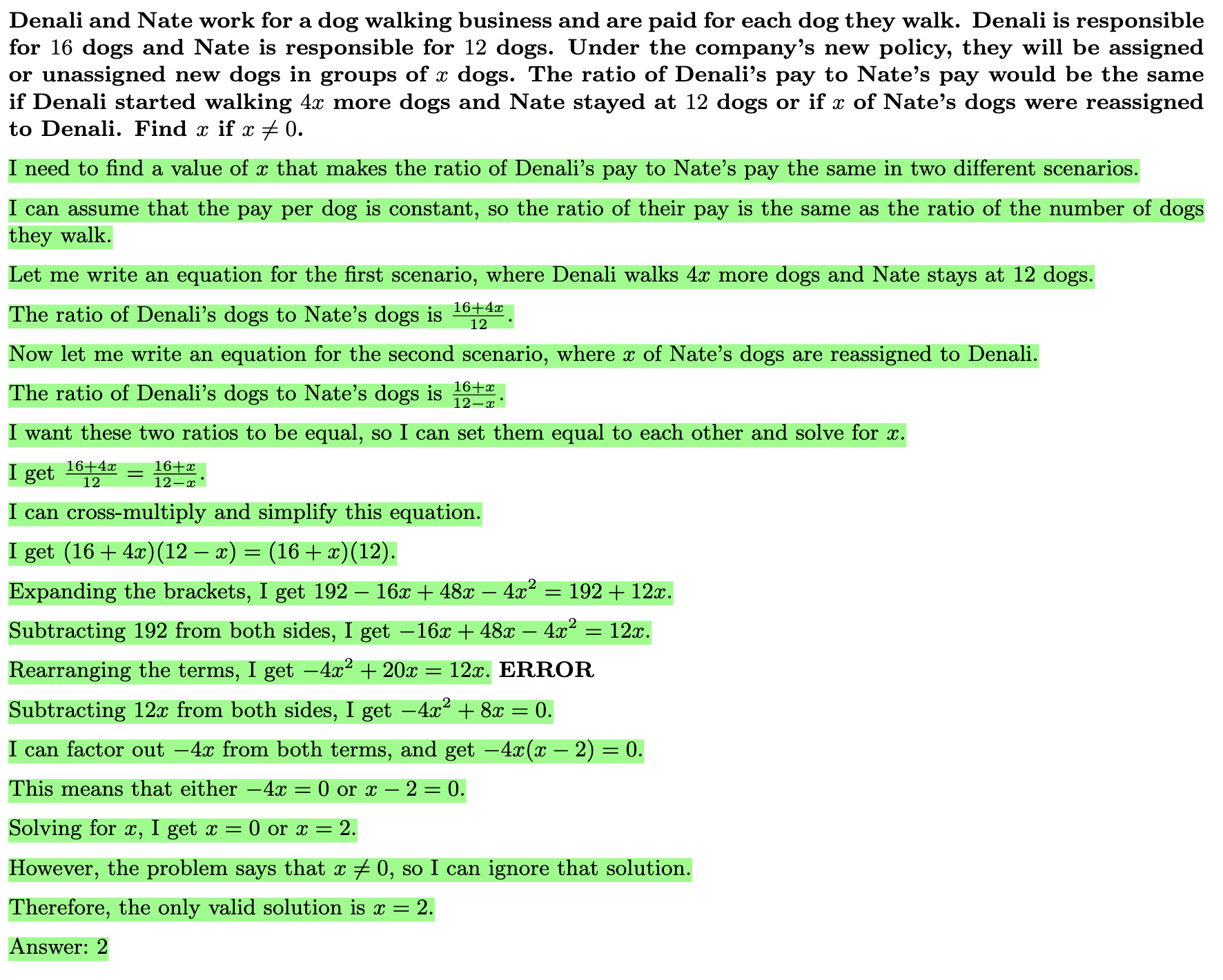}
\end{figure}

\vspace{.5cm}

Problem 11. Generator pass-rate: $13.4\%$. The generator attempts to perform long division, but in step 16, it forgets to include the leading zeros in the repeating part of the decimal. The reward model is fooled by this mistake.

\begin{figure}[!h]
\centering
\includegraphics[width=.7 \textwidth]{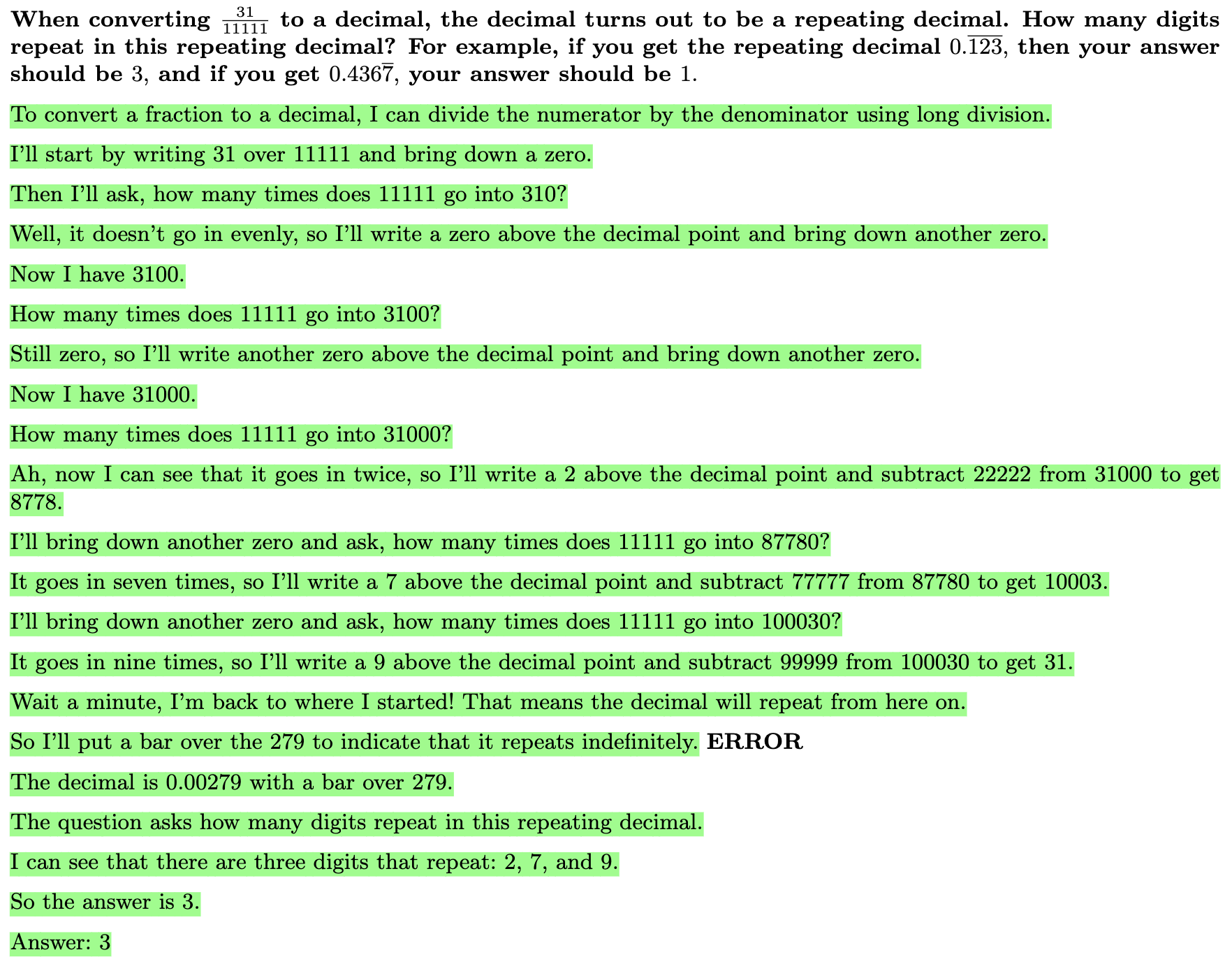}
\end{figure}

\newpage

Problem 12. Generator pass-rate: $9.1\%$. In step 4, the generator falsely claims that the sequence repeats itself every 12 terms, when it’s in fact every 10 terms. This sort of counting mistake occasionally fools the reward model.

\vspace{.25cm}

\begin{figure}[!h]
\centering
\includegraphics[width=.8 \textwidth]{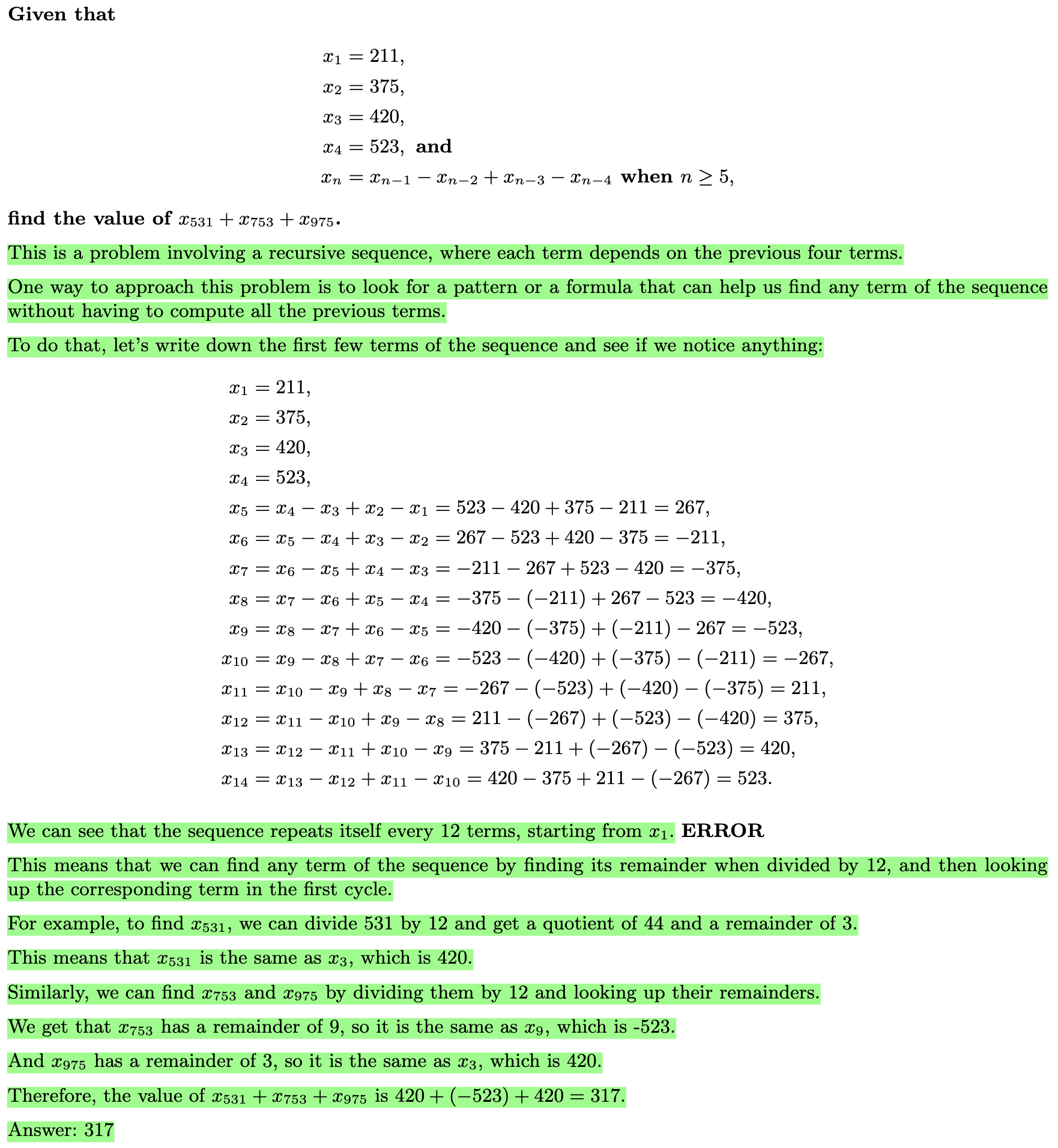}
\end{figure}

\end{document}